\pdfoutput=1
\documentclass[10pt, logo, onecolumn, copyright]{nv}
\usepackage{graphicx}
\definecolor{nvidiagreen}{HTML}{76B900}

\usepackage{mdframed}
\usepackage{color}
\usepackage{xcolor}
\usepackage[table]{xcolor}
\usepackage[utf8]{inputenc}
\usepackage[T1]{fontenc}

\usepackage{amsfonts}
\usepackage{nicefrac}
\usepackage{microtype}
\usepackage{multirow}
\usepackage{multicol}
\usepackage{tabto}
\usepackage{xspace}
\usepackage{amsmath}
\usepackage{adjustbox}
\usepackage{enumitem}
\usepackage{wrapfig}
\usepackage{dblfloatfix}
\usepackage{times}
\usepackage{verbatim}
\usepackage{amssymb}
\usepackage{mathtools}
\usepackage{caption}
\usepackage{subcaption}
\usepackage{array}
\usepackage{colortbl}
\usepackage{booktabs}
\usepackage{bbm}
\usepackage{makecell}
\usepackage{float}
\usepackage{siunitx}
\usepackage{pifont}
\usepackage{marvosym}
\usepackage{listings}
\usepackage{pdflscape}
\usepackage{footmisc}
\usepackage{url}
\usepackage{tabularx}
\usepackage{arydshln}
\usepackage{hhline}
\usepackage{diagbox}
\usepackage{tcolorbox}
\usepackage[nameinlink]{cleveref}
\usepackage{hyperref}
\usepackage[square,sort,comma,numbers]{natbib}
\usepackage{fp}
\usepackage{authblk}
\usepackage{xspace}

\crefname{section}{Sec.}{Sec.}
\crefname{equation}{Eq.}{Eqs.}
\crefname{figure}{Fig.}{Figs.}
\crefname{table}{Tab.}{Tabs.}
\crefname{appendix}{Appendix}{Appendices}
\def\eg{\emph{e.g.}}

\def\vs{\emph{vs.}}

\title{Sol Video Inference Engine: Agent-Native Full-Stack Acceleration Framework for Efficient Video Generation}
\correspondingauthor={\footnotesize \textnormal{Human authors: Yitong Li, Junsong Chen, Haopeng Li, Haozhe Liu, Jincheng Yu, Ligeng Zhu, Ping Luo, Song Han, Enze Xie. 
\\
NVIDIA Research, Efficient AI Team \& Singapore Lab. \\
}}

\author{
\parbox{\linewidth}{
\centering
\vspace{-5pt}
\fontsize{9.5pt}{18pt}\selectfont
NVIDIA SANA Team\textsuperscript{$*$},
Codex$^*$,
GPT-Image2
\vspace{2mm}
\\
\vspace{2pt}
}
}

\begin{abstract}
\vspace{-10pt}

Modern video diffusion models scale toward higher generation quality, but this scaling also increases inference cost.
While a variety of acceleration methods have been proposed, one significant difficulty is that the most effective acceleration implementation is highly \textbf{instance-specific}: the best recipe for one \emph{(model, hardware, inference configuration)} combination often fails to transfer to another.
Models differ in architecture, numerical sensitivity, and attention concentration patterns. Inference configurations vary in spatial and temporal resolution as well as video duration, while hardware platforms differ in memory hierarchy, supported numerics, and kernel throughput.
Together, these factors create a large tuning space in which manual performance engineering becomes costly. To address this, we present \textbf{Sol Video Inference Engine}, an agentic native training-free acceleration framework for video diffusion models. It organizes five broadly applicable techniques---cache, sparse attention, token pruning, quantization, and kernel fusion---into an agentic acceleration stack for instance-specific optimization.
For a concrete \emph{(model, hardware, serving configuration)} deployment target, parallel skill agents optimize the implementation of each technique, an agent integrator composes them into a global stack, and a human validator provides feedback on quality.
We instantiate this workflow on three video models with different model sizes and architectures: 64B \textbf{Cosmos3-Super}, 22B \textbf{LTX-2.3}, and 2B \textbf{SANA-Video}. With little human effort, the full stack achieves more than $2\times$ end-to-end acceleration while maintaining near-lossless VBench quality, demonstrating the effectiveness of the agent framework for video diffusion acceleration.
    \newline
    \textbf{Links:}
    {\hypersetup{urlcolor=nvidiagreen}
    \href{https://github.com/NVlabs/Sana/tree/sol-engine}{Github Code} |
    \href{https://nvlabs.github.io/Sana/Sol-Engine/}{Project Page}
    }
\end{abstract}

\begin{document}

\maketitle

\vspace{1pt}
% Fig. 1 — Hero / head figure (full-width teaser)
\begin{figure}[H]
\centering
\includegraphics[width=0.94\linewidth]{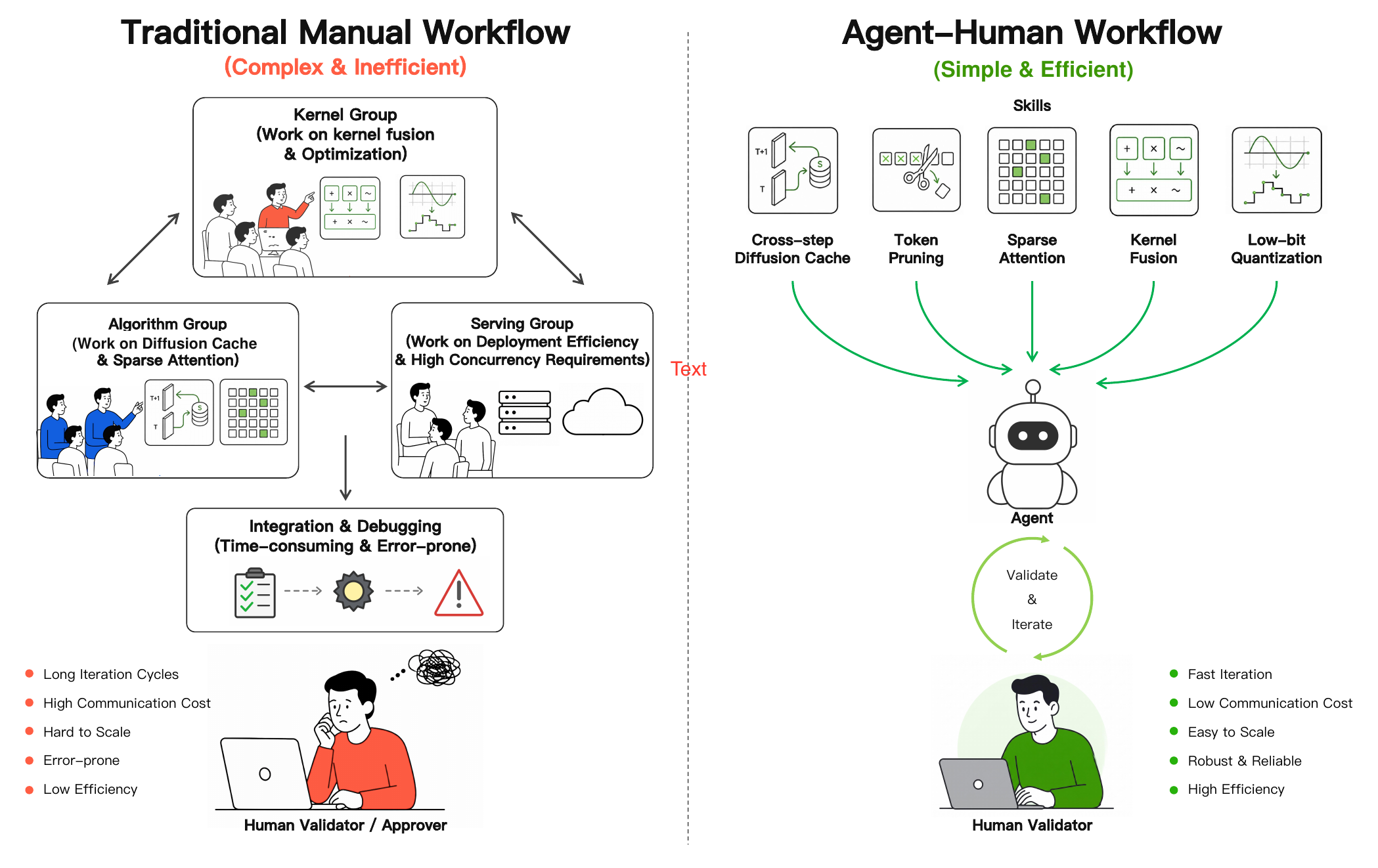}
\caption{\textbf{Traditional manual acceleration workflow versus Sol Video Inference Engine's agent-human workflow.} Left: acceleration work is split across multiple human groups with different expert techniques, leading to long iteration cycles, high communication cost, and manual integration or debugging effort. Right: Sol Video Inference Engine exposes acceleration skills to an agent system that performs local search and stack integration, while the human stays in the loop mainly for validation and iteration. This agent-human mode reduces coordination overhead and human effort while making the workflow faster, easier to scale, and more reliable.}
\label{fig:teaser}
\end{figure}

\section{Introduction}
\label{sec:intro}

Video diffusion models continue to scale toward higher generation quality, longer duration, and higher fidelity.
This scaling, however, also increases inference cost.
Modern video diffusion systems combine billion-scale backbones, long spatiotemporal sequences, and multi-step sampling, making inference computationally heavy.

While many acceleration methods have been proposed, accelerating these systems is not a one-time optimization problem.
One significant difficulty is that effective acceleration is highly \textbf{instance-specific}: the best recipe for one \emph{(model, hardware, inference serving configuration)} combination often fails to transfer to another.
Models differ in architecture, numerical sensitivity, and attention concentration patterns.
Inference configurations vary in spatial and temporal resolution, video duration, denoising schedule, and batching strategy, while target hardware ranges from high-memory datacenter cards to consumer-grade edge devices with distinct supported numerics and throughput.

As a direct result of these heterogeneous properties, the implementation of various acceleration methods is instance-dependent: a cache policy may be valid for one denoising schedule but drift under another; a sparsity pattern may preserve quality at one resolution but fail at a longer temporal context; and a low-precision backend may provide a good speed--quality trade-off for one hardware but offer a poor trade-off for another.
Because the combination space spanned by these factors is so large, hand-tuning every deployment instance consumes substantial engineering effort.

Sol Video Inference Engine addresses this challenge with an \textbf{agentic acceleration framework} organized around two coupled parts.
The first is a full-stack set of acceleration techniques spanning cross-step cache, sparse attention, token pruning, quantization, and kernel fusion.
The second is an agent-native workflow: parallel skill agents perform per-technique local optimization, an agent integrator composes the resulting candidates through global search, and a human validator provides visual-quality feedback.
Sol Video Inference Engine applies this framework to 64B Cosmos3-Super, 22B LTX-2.3, and 2B SANA-Video, producing effective acceleration stacks with little manual effort. The contributions of this paper can be summarized in four aspects:
\begin{itemize}[leftmargin=1.5em, itemsep=2pt]
    \item \textbf{Problem reformulation.} We systematically analyze the acceleration space of video diffusion inference and formulate its acceleration as an instance-specific tuning problem rather than one-time optimization paradigms.
    
    \item \textbf{Full-stack acceleration framework.} We propose a comprehensive framework for video diffusion models acceleration, synergistically integrating algorithmic and system-level techniques, including cross-step caching, sparse attention, token pruning, quantization, and kernel fusion.
    
    \item \textbf{Agent-driven optimization workflow.} We propose an agent-native architecture that orchestrates parallel agents and an integrator. It successfully bypasses the multi-team engineering traditionally required for model deployment, autonomously delivering highly optimized speed--quality configurations in a fraction of the usual human effort.
    
    \item \textbf{Extensive empirical validation.} Extensive experiments demonstrate that our training-free framework achieves over $2\times$ end-to-end speedup across diverse state-of-the-art models—including 64B Cosmos3-Super, 22B LTX-2.3, and 2B SANA-Video—without compromising visual quality.
    \end{itemize}

\paragraph{Report organization.}
Section~\ref{sec:related} reviews related work.
Section~\ref{sec:background} analyzes the deployment constraints and instance-specific nature of video diffusion acceleration.
Section~\ref{sec:overview} presents the agentic acceleration framework, the local-to-global workflow, and the acceleration techniques used in that process.
Section~\ref{sec:experiments} evaluates Sol Video Inference Engine on Cosmos3-Super, LTX-2.3, and SANA-Video.
Section~\ref{sec:conclusion} concludes.

\section{Related Work}
\label{sec:related}

\subsection{Video Diffusion Models}
\label{subsec:related_video}

Video generation is currently dominated by large-scale Diffusion Transformers (DiTs). Mainstream open-source models, such as CogVideo and CogVideoX~\cite{hong2022cogvideo,yang2025cogvideox}, the Wan series~\cite{wan2025}, HunyuanVideo~\cite{kong2024hunyuanvideo,hunyuanvideo2025}, and Cosmos3-Super~\cite{nvidia2026cosmos3super}, establish the standard paradigm for high-fidelity and physically-aware synthesis. To handle extended temporal contexts and high resolutions, systems like LongCat-Video~\cite{meituan2025longcatvideo}, LTX-2.3~\cite{lightricks2026ltx23}, JoyAI-Echo~\cite{echo2026longvideo}, and Pyramid Flow~\cite{jin2024pyramidal} adopt multi-stage, autoregressive, or memory-augmented generation pipelines. Other models explore the extremes of the design space for efficiency: SANA-Video~\cite{chen2025sana} prioritizes efficiency through linear attention. Despite this architectural diversity, all these systems face a severe, shared inference bottleneck. The combination of massive spatiotemporal token sequences and iterative denoising steps inherently makes efficient deployment a critical system-level challenge.

\subsection{Video Generation Acceleration}
\label{subsec:related_acceleration}

Prior acceleration work attacks video diffusion cost from three complementary levels.
At the \textbf{algorithm} level, cache-based methods exploit similarity across denoising steps: TeaCache predicts timestep-to-timestep output change to decide when cached residuals can be reused~\cite{liu2024timestep}, and TaylorSeer forecasts features across timesteps using Taylor expansion to reduce the error of more aggressive reuse~\cite{liu2025taylorseer}. A broader spectrum of caching strategies has also emerged to accelerate diffusion inference~\cite{zhou2025easycache,cachedit2025,zhao2024pab}.
At the \textbf{model} level, sparse attention and token reduction reduce work inside each denoising step: PISA uses training-free piecewise sparse attention with first-order Taylor compensation~\cite{li2026pisa}, Sparse VideoGen and Sparse VideoGen2 exploit spatial-temporal sparsity with profiling and layout-aware sparse execution~\cite{xi2025sparse,yang2025sparse}, VSA learns video sparse attention operators~\cite{zhang2025vsa}. Many other methods have also extensively explored sparse attention for video diffusion~\cite{zhang2025spargeattn,shen2025draft,zhang2025sta,xu2025xattention,li2025radial,chen2026longlive20}. Meanwhile, token-reduction methods such as ToMe-SD, Astraea, TAPE, and CoReDiT merge, prune, or reconstruct redundant diffusion tokens~\cite{bolya2023tomesd,liu2025astraea,li2026tape,li2026coredit}.
At the \textbf{kernel} level, quantization and fusion reduce the cost of the remaining operators: PTQ4DiT, Q-DiT, ViDiT-Q, and SVDQuant study diffusion-aware low-precision execution~\cite{wu2024ptq4dit,chen2025qdit,zhao2025viditq,li2025svdquant,li2026fp4explorebf16train}. Moreover, CUTLASS epilogues, ByteTransformer, and CODA show how GEMM-adjacent memory-bound operations can be fused into high-throughput kernels~\cite{cutlassEpilogue,zhai2023bytetransformer,guo2026coda}.
Sol Video Inference Engine builds on these technique families.

\subsection{Agentic Workflows and Harness Engineering}
\label{subsec:related_agentic}

Recent work increasingly treats the surrounding agent runtime---tool access, execution loop, verification, memory, and feedback---as a first-class optimization target rather than incidental glue code.
AgentBench and MLAgentBench provide broad evaluation settings for language agents in interactive environments and machine-learning experimentation, respectively~\cite{liu2024agentbench,huang2024mlagentbench}.
In software engineering, SWE-agent, AutoCodeRover, Agentless, and OpenHands study how agent interfaces, search, localization, execution environments, and generalist developer agents affect autonomous code improvement workflows~\cite{yang2024sweagent,zhang2024autocoderover,xia2024agentless,wang2024openhands}.
AI Harness Engineering formalizes the harness itself as a runtime substrate for foundation-model software agents~\cite{zhong2026ahe}, while The AI Scientist shows that agentic systems can orchestrate idea generation, experiment execution, figure production, and manuscript writing in an end-to-end loop~\cite{lu2024aiscientist}.
Closer to systems optimization, CUDA-LLM and CudaForge demonstrate that iterative agentic workflows with compilation checks, correctness validation, profiling, and hardware feedback can automatically generate or optimize efficient CUDA kernels~\cite{chen2025cudallm,zhang2025cudaforge}.
Sol Video Inference Engine is aligned with this direction, but targets a different object of coordinating multiple acceleration skills for video diffusion inference into a quality-constrained full-stack deployment.

\section{Background: Inference Redundancy and Deployment Heterogeneity}
\label{sec:background}

\subsection{Inference Redundancy in Video Diffusion Models}
\label{subsec:video_diffusion}
\label{subsec:redundancy_space}

\begin{figure}[t]
\centering
\begin{subfigure}[t]{0.42\linewidth}
\centering
\includegraphics[width=\linewidth]{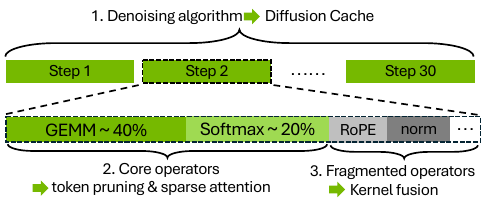}
\caption{Three-level redundancy.}
\label{fig:baseline_time_breakdown_levels}
\end{subfigure}
\hfill
\begin{subfigure}[t]{0.54\linewidth}
\centering
\includegraphics[width=\linewidth]{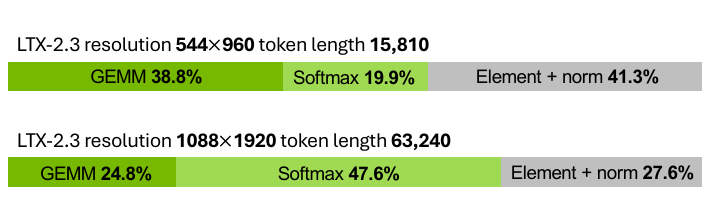}
\caption{Time profile across diffusion configurations under the same model and hardware.}
\label{fig:ltx_diffusion_config_time_profile}
\end{subfigure}
\caption{\textbf{Baseline inference time breakdown and configuration-specific bottlenecks.} Left: runtime decomposition that illustrates algorithm-, model-, and kernel-level redundancy in the unaccelerated serving path. Right: time profile across diffusion configurations under the same model and hardware. Different spatial resolutions shift the dominant inference bottlenecks and therefore expose different primary optimization opportunities.}
\label{fig:baseline_time_breakdown}
\end{figure}

Scaling video diffusion models improves generation quality, temporal consistency, and controllability, but it also substantially increases inference cost.
Recent systems illustrate this trend: Cosmos3-Super scales to 64B parameters for high-capacity video and physical-aware content generation~\cite{nvidia2026cosmos3super}, while LTX-2.3 is released as a 22B-parameter audio-video diffusion foundation model~\cite{lightricks2026ltx23}.
As parameter count and model scale increase, video diffusion inference also exposes redundancy at three complementary levels.
At the \textbf{diffusion algorithm} level, adjacent denoising steps often execute structurally similar computations over slowly changing latent states, creating opportunities for cross-step reuse, selective skipping, and compensation.
At the \textbf{model} level, long spatiotemporal sequences contain redundant tokens and attention interactions, so the full attention graph or full token set is not always necessary at every layer and timestep.
At the \textbf{kernel} level, transformer inference repeatedly launches memory-bound operators around GEMMs, materializes intermediate tensors, and pays overhead for layout movement, normalization, activation, and precision conversion.
Figure~\ref{fig:baseline_time_breakdown_levels} visualizes how the three levels of redundancy appear in the unaccelerated inference path.

% Fig. 2 — Baseline time breakdown and configuration-specific bottlenecks

\subsection{Deployment Heterogeneity: Model, Hardware, and Configuration}
\label{subsec:serving_config}
\label{subsec:non_transferable}

\begin{figure}[t]
\centering
\includegraphics[width=0.95\linewidth]{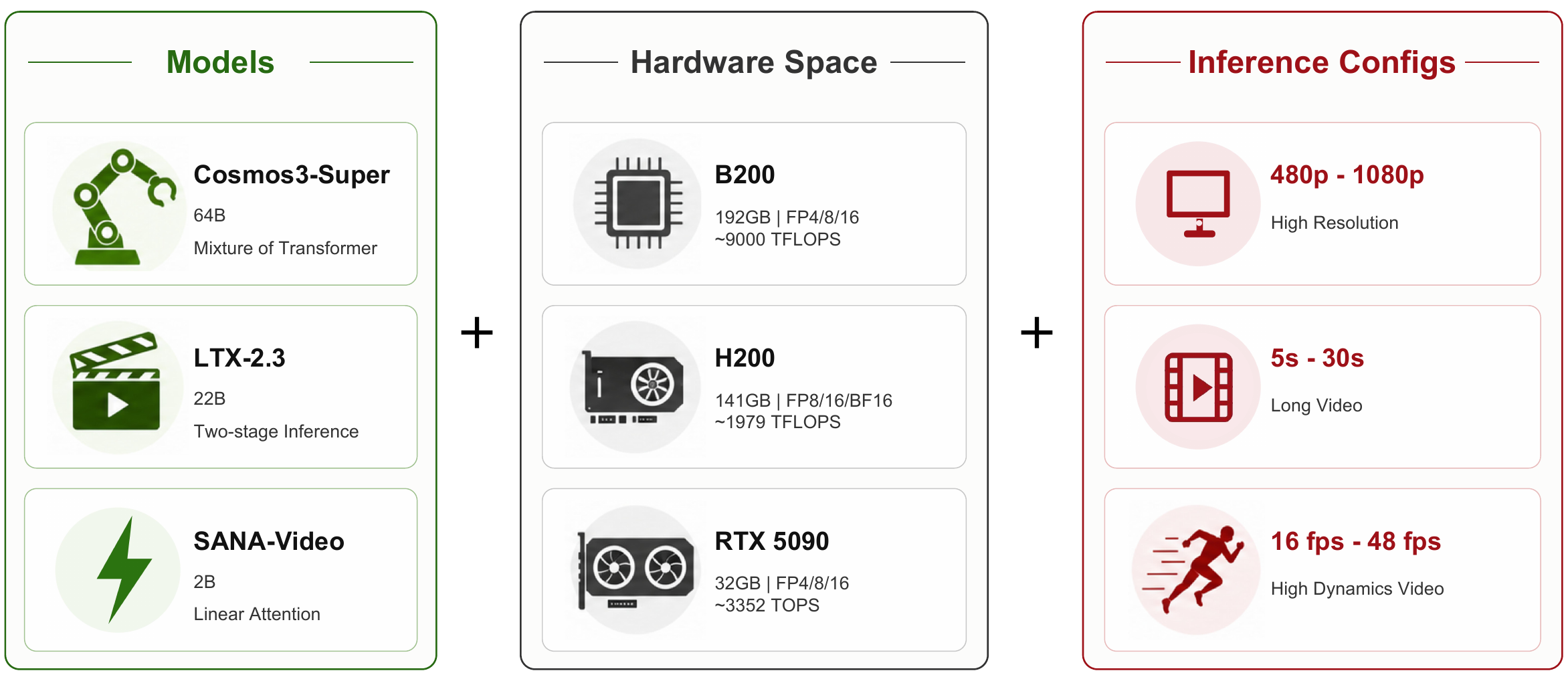}
\caption{\textbf{Joint combination space of video generation model deployment instances.} Model, hardware, and inference configuration jointly define a large combinatorial space of deployment instances. This diversity makes implementation of acceleration techniques a substantial engineering challenge rather than a one-time optimization problem.}
\label{fig:combination_space}
\end{figure}

We treat an acceleration target as a heterogeneous deployment instance defined by three co-equal axes: \emph{model, hardware, and inference serving configuration}. Acceleration implementations fail to transfer cleanly because each deployment instance is constrained by the joint choice of them, spanning a large combination space as shown in Figure~\ref{fig:combination_space}.
The \textbf{model} determines numerical and structural sensitivity: different backbones have different quantization robustness, attention redundancy, cache stability across denoising steps, and tolerance to token pruning.
The \textbf{hardware} determines which bottlenecks are exposed: on high-compute GPUs such as B200, large GEMMs may become sufficiently fast that small-operator launch overhead and memory-bound epilogues occupy a larger fraction of runtime, while on lower-compute GPUs those same launches may be less visible because dense compute remains dominant. 
The \textbf{configuration} axis covers runtime choices such as spatial and temporal resolution (\eg 720p \vs 1080p, 24fps \vs 48 fps). It changes the effective token count and runtime balance: increasing spatial or temporal resolution increases sequence length, attention scales quadratically with token count, and FFN/MLP work scales roughly linearly, so the attention--FFN time split can change substantially across resolutions.
These coupled constraints make caching, sparsity, pruning, fusion, and quantization properties instance-specific; an acceleration recipe that works for one \emph{(model, hardware, serving configuration)} point may not preserve speed or quality at another.
Figure~\ref{fig:ltx_diffusion_config_time_profile} illustrates this effect within a single model and hardware setting: different resolutions expose different dominant bottlenecks and therefore different primary acceleration directions.

This creates two consequences for performance engineering.
First, the number of deployment instances is large, so exhaustive manual tuning does not scale.
Second, the optimization problem within each instance is structured: objectives such as latency and quality-preservation are explicit, which naturally lends itself to agent-based automation.
Sol Video Inference Engine on Cosmos3-Super, LTX-2.3, and SANA-Video instantiates distinct points in this configuration space, demonstrating how an agent-native workflow can produce effective stacks for concrete deployment instances.

\section{Agentic Acceleration Stack}
\label{sec:overview}
\label{sec:stack}

\begin{figure}[t]
\centering
\includegraphics[width=\linewidth]{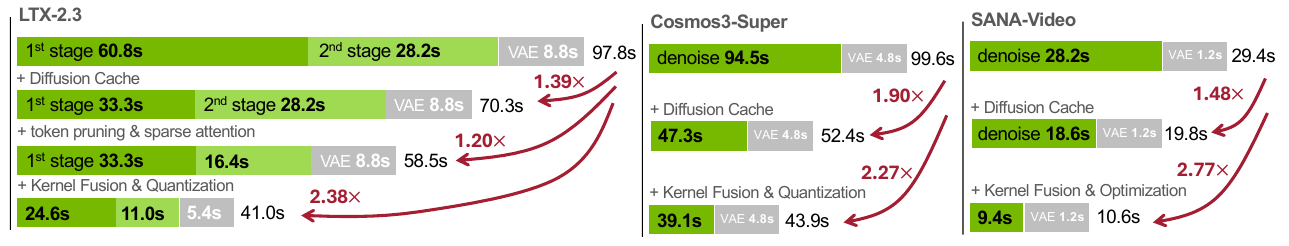}
\caption{\textbf{Acceleration breakdown across different optimization methods.} The figure decomposes the cumulative speedup contributed by different components in the Sol Video Inference Engine stack, showing how cache, sparse attention and token pruning, and kernel-level optimization progressively reduce end-to-end latency. Overall, the full framework delivers roughly $2\times$--$3\times$ acceleration across Cosmos3-Super, LTX-2.3, and SANA-Video.}
\label{fig:time_profile}
\end{figure}

Sol Video Inference Engine builds a full-stack acceleration framework that spans three levels of acceleration: algorithm-level cache to reduce redundancy across repeated denoising steps, model-level sparse attention and token pruning to reduce unnecessary token computation, and kernel-level optimizations through kernel fusion to minimize kernel-launch overhead, alongside quantization to accelerate execution by leveraging lower-precision hardware primitives.
For a specific deployment instance, defined by a concrete \emph{(model, hardware, serving configuration)} combination, the framework aims to select and compose the most suitable subset of these acceleration skills under acceptable generation quality. Within this framework, each acceleration technique is first tuned locally to identify the configuration that best fits the current deployment instance.
The agent integrator then composes these per-technique optima and performs global tuning to find the strongest full-stack combination under generation-quality constraints.
Sol Video Inference Engine delivers roughly $2\times$--$3\times$ end-to-end acceleration across Cosmos3-Super, LTX-2.3, and SANA-Video, which is achieved by a full-stack techniques composition rather than a single acceleration method, as shown in Figure~\ref{fig:time_profile}.

% Fig. 4 — Acceleration breakdown across optimization methods

% Fig. 5 — Agentic acceleration workflow / stack
\begin{figure}[t]
\centering
\includegraphics[width=\linewidth]{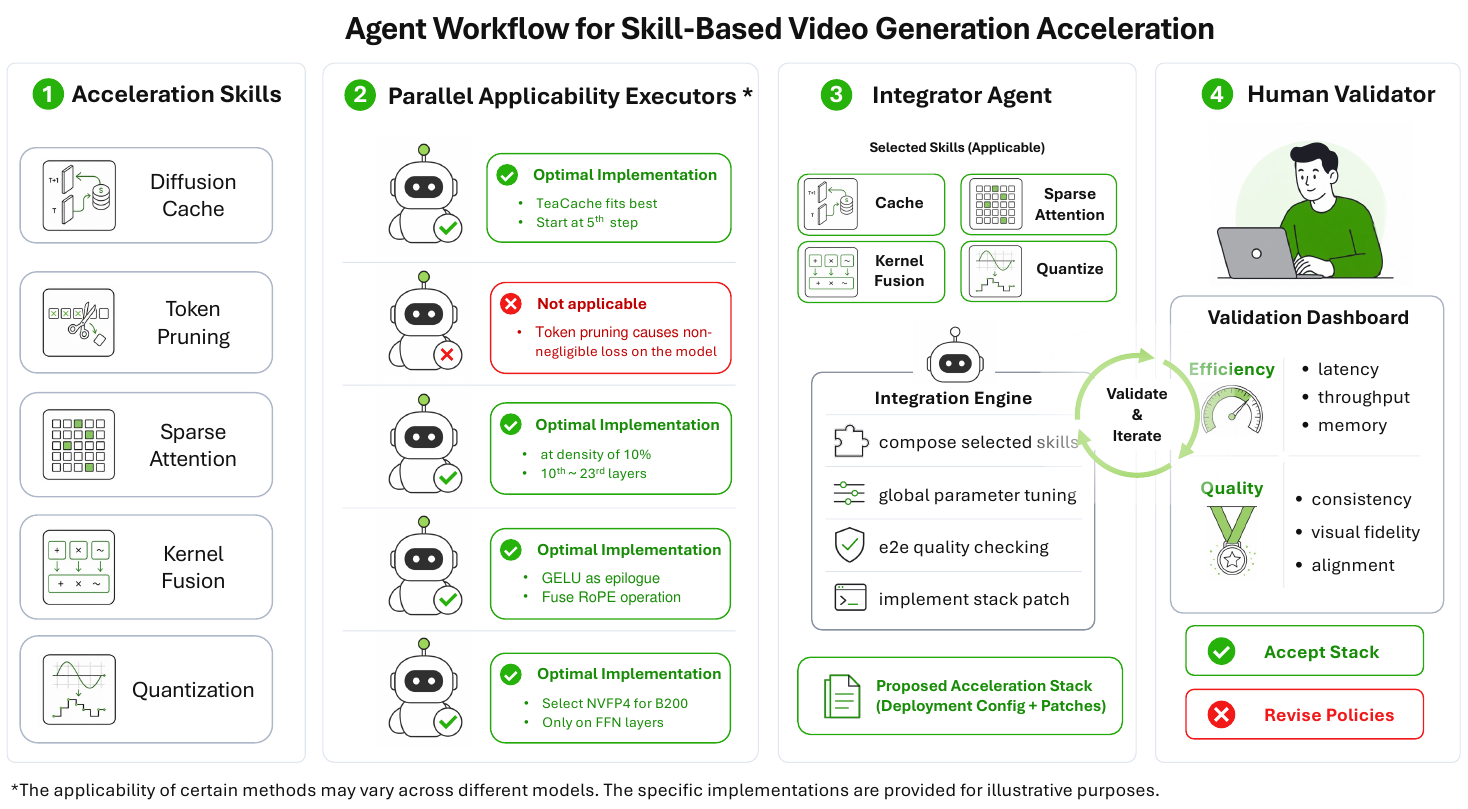}
\caption{\textbf{Agent-native acceleration workflow.} The workflow can be viewed in two parts. The first part is local parallel per-technique search, where skill agents independently explore local candidates. The second part is global integration and the human-validator loop, where an integrator composes the local candidates into a full acceleration stack and the human validator provides feedback to guide the next iteration.}
\label{fig:accel_skills}
\end{figure}

\subsection{Agent-Native Acceleration Workflow}
\label{sec:workflow}
\label{subsec:agent_workflow}

Sol Video Inference Engine uses an \textbf{agent-native acceleration workflow} to tune and compose a deployment stack for each instance.
The workflow turns the technique families detailed in Section~\ref{subsec:acceleration_stack} into a structured optimization process.
Illustrated in  Figure~\ref{fig:accel_skills}, it proceeds in three stages: parallel skill agents perform model-specific local optimization, an agent integrator composes the local candidates into a global stack, and a human validator provides quality feedback.

\begin{figure}[t]
\centering
\includegraphics[width=\linewidth]{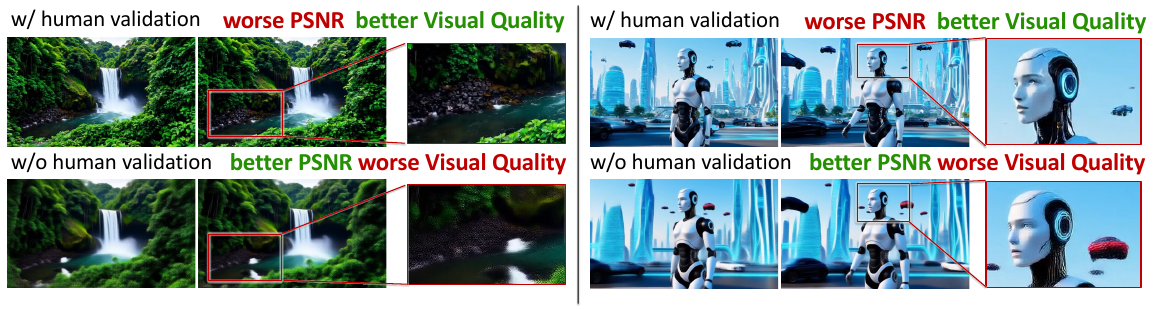}
\caption{\textbf{Necessity of human validation.} Human validators inspect representative generations and provide feedback on overall visual quality, while traditional similarity operators such as PSNR often fail to faithfully reflect the actual perceptual quality of generated videos in deployment-oriented evaluation.}
\label{fig:human_verify}
\end{figure}

\begin{itemize}[leftmargin=1.5em, itemsep=2pt]
    \item \textbf{Parallel skill agents.} We first perform local optimization within each technique family. Because the full stack contains many techniques and the joint search space is large, directly running a single agent over the entire global optimization problem would be costly and slow. Instead, independent skill agents search their own local spaces in parallel. For example, a cache agent searches skip-step schedules and compensation rules, while a kernel-fusion agent profiles the kernel-wise time cost and tests fusions such as GEMM + GELU or RoPE + norm. These locally optimized candidates provide strong per-skill starting points for the later global optimization stage.
    \item \textbf{Agent as integrator.} The integrator receives local candidates from all skill agents and combines them into a full acceleration recipe for the target deployment. Because these methods are not independent, the final end-to-end speedup is not the direct product of their local gains. Moreover, many techniques are lossy, so directly stacking the locally optimal choices can accumulate approximation error and produce an unacceptable final result. The integrator therefore performs global optimization rather than simply aggregating first-stage results. It resolves interactions such as whether lossy methods like cache and sparse attention remain acceptable when combined, and whether quantization admits a compatible implementation together with kernel fusion under the same deployment target.
    \item \textbf{Human validator feedback.} Similarity metrics such as PSNR are not well aligned with human visual perception. Given a high-quality reference video, PSNR may assign a large distance to outputs with shifted visual details, while assigning a smaller distance to outputs with blur, temporal jitter, or degraded motion coherence, as shown in Figure~\ref{fig:human_verify}. From a human-quality perspective, the former is acceptable while the latter is not, which makes human feedback necessary. In the workflow, we use a fixed validation set to inspect each integrated candidate and determine whether its visual quality remains acceptable. This feedback then tells the next optimization round whether the global agent can push toward more aggressive acceleration or should stay more conservative to preserve quality.
\end{itemize}

% Fig. 6 — Human feedback validation in the agent-native workflow

It seeks strong local configurations for each technique family and composes them into a global operating point under quality constraints.
In this process, parallel agents optimize each technique family locally, the integrator searches over the resulting candidate points, and the human validator verifies visual quality.
Together, this workflow enables instance-specific optimization for each deployment target without requiring large manual tuning effort.

\subsection{Acceleration Techniques}
\label{subsec:acceleration_stack}

\subsubsection{Cross-Step Cache}
\label{subsec:algorithm_level}
\label{subsec:cache}
\label{subsec:cache_detail}

Diffusion inference typically requires a large number of function evaluations (NFE), often on the order of 30--50 denoising steps for high-quality generation.
Across these steps, the model repeatedly executes structurally similar computation, and adjacent denoising states often induce similar intermediate features.
Cross-step caching targets this trajectory-level redundancy by skipping selected denoising-step computations and compensating for the skipped output.

Recent diffusion-cache methods illustrate the range of usable cache policies.
TeaCache estimates timestep-to-timestep output change from timestep-conditioned inputs and reuses cached residuals when the predicted change is small~\cite{liu2024timestep}.
EasyCache introduces a runtime-adaptive caching policy that adjusts feature reuse online to better balance acceleration and quality across video diffusion trajectories~\cite{zhou2025easycache}.
TaylorSeer extends feature reuse into feature forecasting, using Taylor-series extrapolation over previous timestep features to reduce quality degradation when cached intervals become larger~\cite{liu2025taylorseer}.
For a specific model and deployment configuration, Sol Video Inference Engine treats cache selection as an instance-specific optimization problem rather than a generalizable methodology. 
The cache agent must determine which policy best matches the model, then tune its key hyperparameters, such as skip-step schedule, cached feature or residual choice, compensation rule, warmup, and maximum cached-step constraints.

\subsubsection{Sparse Attention}
\label{subsec:model_level}
\label{subsec:compute_reduction}
\label{subsec:sparse_attention}

Video diffusion inference operates over large spatiotemporal token sets, especially for long-duration or high-resolution generation.
These tokens exhibit both temporal redundancy across adjacent frames and spatial redundancy within each frame, so many attention edges contribute little to the final output on a given step.
Sparse attention targets the quadratic cost of full spatiotemporal attention without necessarily changing the external token lattice.

Recent work shows several ways to exploit this sparsity.
PISA provides training-free piecewise sparse attention: critical blocks are computed exactly, while non-critical blocks are compensated with first-order Taylor approximation~\cite{li2026pisa}.
SpargeAttention introduces a universal training-free sparse attention framework that accurately predicts and skips near-zero attention entries using a block-wise similarity metric~\cite{zhang2025spargeattn}.
Furthermore, Sparse VideoGen exploits spatial-temporal sparsity patterns with online profiling and custom kernels~\cite{xi2025sparse}, while Sparse VideoGen2 improves sparse-token identification and GPU layout through semantic-aware permutation~\cite{yang2025sparse}.
For a specific model and deployment configuration, Sol Video Inference Engine must determine which sparse attention backend fits best in the setting, as well as which attention layers can be sparsified safely and which should remain dense for quality preservation.
The sparse-attention agent therefore tunes layer selection, sparsity pattern, and compensation policy, then validates the quality--speedup trade--off under given inference configuration.

\subsubsection{Token Pruning}
\label{subsec:token_pruning}

Token pruning reduces the sequence itself by skipping, merging, or reconstructing redundant latent tokens.
By reducing the number of active tokens, it lowers the cost of subsequent attention and feed-forward computation, especially in high-resolution or long-video settings where sequence length dominates latency.

Earlier token-reduction work such as ToMe for Stable Diffusion shows that redundant diffusion tokens can be merged at inference time with limited quality loss~\cite{bolya2023tomesd}.
More recent video-diffusion systems make the pruning decision more temporally and spatially aware: Astraea searches token budgets for video DiTs under a performance target~\cite{liu2025astraea}, TAPE smooths token importance across frames to avoid temporal jitter~\cite{li2026tape}, and CoReDiT prunes spatially coherent DiT tokens while reconstructing skipped outputs to preserve a dense representation~\cite{li2026coredit}.
For a specific model and deployment configuration, the key question is not whether token pruning helps in principle, but which pruning policy removes redundant computation without destabilizing motion or fine detail.
The token-pruning agent therefore tunes the pruning criterion, pruning ratio, layer schedule, timestep schedule, and reconstruction rule, then validates each candidate against temporal coherence and late-step fidelity before exposing it to stack integration.

\subsubsection{Quantization}
\label{subsec:quant}

While quantization is typically employed to reduce computational cost, its actual speedup is heavily governed specifically by the model's hidden dimension, the number of inference tokens and the hardware characteristics. Concurrently, a model's sensitivity to quantization noise is fundamentally dictated by its inherent training parameter distribution. Because both the acceleration potential and the resulting quality impact vary drastically across various deployment instances, we must conduct an instance-specific search to identify the optimal quality–speedup trade-off.

Recent work reduces quantization loss through diffusion-aware calibration and scaling.
PTQ4DiT identifies salient channels and timestep-varying activations as key post-training quantization challenges~\cite{wu2024ptq4dit}; Q-DiT uses automatic quantization granularity allocation for DiTs~\cite{chen2025qdit}; and SVDQuant absorbs weight and activation outliers with low-rank components to enable 4-bit diffusion inference~\cite{li2025svdquant}.
The SageAttention series targets the attention quantization, introducing 8-bit to 4-bit attention acceleration with outlier smoothing~\cite{zhang2025sageattention,zhang2025sageattention2,zhang2025sageattention3}.
For a specific deployment instance, Sol Video Inference Engine treats quantization as a selective local-tuning problem rather than a uniform low-bit conversion.
The quantization agent profiles layer sensitivity and tensor shapes, then tunes layer-wise precision assignments, weight and activation bitwidths, timestep-dependent scaling to find the best speed--quality trade-off.

\subsubsection{Kernel Fusion}
\label{subsec:kernel_level}
\label{subsec:operator_efficiency}
\label{subsec:fusion}

Transformer and diffusion backbones spend substantial time around GEMMs on fragmented operators that have low arithmetic intensity: bias addition, residual updates, normalization, activation functions, scaling, and layout conversion.
Kernel fusion accelerates these paths by executing memory-bound follow-up operators while intermediate results are still in registers or shared memory, instead of materializing them to HBM and launching separate kernels.

This pattern is explicit in CUTLASS epilogue programming, where activations such as GELU can be fused as GEMM epilogues~\cite{cutlassEpilogue}, and in ByteTransformer, which uses customized fused CUTLASS epilogues for bias and GELU to hide memory latency inside GEMM execution~\cite{zhai2023bytetransformer}.
CODA generalizes this idea by rewriting transformer blocks as GEMM-plus-epilogue programs, fusing normalization, activations, residual updates, reductions, and related memory-bound work into tile-local epilogue code~\cite{guo2026coda}.
For a specific model and deployment configuration, Sol Video Inference Engine treats kernel fusion as a late-stage local-tuning problem driven by the current bottlenecks of the partially accelerated stack.
The kernel-fusion agent profiles the kernel-wise time consumption, then tunes which operator sequences to fuse, whether to use custom kernels or compiler-generated fusion, and how to match the implementation to the current tensor shapes and precision formats.
Typical candidates include GEMM + GELU, GEMM + residual, and fused RoPE + norm.

\section{Experiments}
\label{sec:experiments}

Sol Video Inference Engine applies the agent-native workflow to multiple video diffusion models.
Each deployment is a distinct \emph{(model, hardware, inference serving configuration)} instance: the deployments share the same technique stack, but concrete implementations must be applied, optimized, and validated per target.
We evaluate three representative video models across different model sizes and architectures---Cosmos3-Super (64B)~\cite{nvidia2026cosmos3super}, LTX-2.3 (22B)~\cite{lightricks2026ltx23}, and SANA-Video (2B)~\cite{chen2025sana}---to demonstrate the generalizable acceleration capability of our framework.

\subsection{Experimental Protocol}
\label{subsec:setup}

\paragraph{Deployment Hardware.} All experiments run on NVIDIA B200 GPUs. We first deploy each model in an SGLang-based serving stack and then develop the acceleration components on top of this deployment. Cosmos3-Super uses four GPUs with sequence parallelism (SP), while LTX-2.3 and SANA-Video use one GPU.

% \textbf{Baselines.} We compare against the unaccelerated serving path and apply the optimized stack in diffusion inference frameworks built around SGLang~\cite{zheng2024sglang}, vLLM~\cite{kwon2023vllm}, and Cache-DiT~\cite{cachedit2025}. In these baselines, SGLang and vLLM provide serving/runtime infrastructure without embedding diffusion-specific acceleration techniques, while Cache-DiT includes cache as a single acceleration method.

\paragraph{Metrics.} We use end-to-end latency as the primary metric. For quality preservation, we use VBench~\cite{huang2024vbench} to evaluate visual quality and motion quality under matched prompts and serving configurations.

\paragraph{Models.}
We choose Cosmos3-Super~\cite{nvidia2026cosmos3super}, LTX-2.3~\cite{lightricks2026ltx23}, and SANA-Video~\cite{chen2025sana} to evaluate acceleration under our agentic acceleration framework.
These models span different parameter scales, from 2B to 64B, and cover distinct inference pipelines and architectural designs: Cosmos3-Super uses a Mixture-of-Transformers (MoT) structure for physical video generation, LTX-2.3 emphasizes multi-resolution two-stage inference, and SANA-Video replaces quadratic attention with linear attention.
Accelerating all three models demonstrates that the framework can generalize across foundation models with varied scales and architectural designs rather than overfitting to a single model family.
\begin{itemize}[leftmargin=1.5em, itemsep=2pt]
    \item \textbf{Cosmos3-Super.} Cosmos3-Super is a 64B-parameter large video model focused on physical generation~\cite{nvidia2026cosmos3super}. It uses a Mixture-of-Transformers (MoT) architecture with an autoregressive reasoning tower and a diffusion generation tower, scaling the model to 64B parameters and improving generation quality and physical reasoning capability.
    \item \textbf{LTX-2.3.} LTX-2.3 is a 22B-parameter audio-video diffusion model~\cite{lightricks2026ltx23}. Its serving pipeline includes two stages: stage~1 generates the video latent at $544{\times}960$ with 15 denoising steps, stage~2 upsamples and refines at $1088{\times}1920$ with 3 denoising steps, followed by VAE decode for 241 frames. The high-quality pipeline uses the \texttt{res\_2s} second-order sampler, which improves quality/step trade-offs but differs from the Euler-style schedules assumed by many cache heuristics, making cache implementation and optimization less direct.
    \item \textbf{SANA-Video.} SANA-Video is a 2B-parameter video diffusion model based on a block linear diffusion transformer~\cite{chen2025sana}. It uses linear attention to reduce the $O(n^2)$ overhead of standard attention and convolution-enhanced FFN blocks to preserve local spatiotemporal structure, achieving strong video quality while remaining efficient.
\end{itemize}

\subsection{Efficiency Analysis}
\label{subsec:results_ltx}
\label{subsec:e2e_ltx}

\begin{table}[t]
\centering
\caption{\textbf{End-to-end acceleration across model deployments on NVIDIA B200 GPUs.} The table compares baseline inference systems against the fully optimized Sol-Engine configuration across Cosmos3-Super, LTX-2.3, and SANA-Video. Speedups are reported relative to the SGLang baseline, while official latencies are shown for reference. The final optimized system delivers roughly $2\times$--$3\times$ end-to-end speedup.}
\label{tab:ablation_stack}
\scriptsize
\resizebox{0.78\textwidth}{!}{%
\begin{tabular}{lcccccc}
\toprule
\multirow{2}{*}{\textbf{Framework}} & \multicolumn{2}{c}{\textbf{Cosmos3-Super (64B, GPU=4)}} & \multicolumn{2}{c}{\textbf{LTX-2.3 (22B)}} & \multicolumn{2}{c}{\textbf{SANA-Video (2B)}} \\
\cmidrule(lr){2-3} \cmidrule(lr){4-5} \cmidrule(lr){6-7}
& {\footnotesize Latency (s)} & {\footnotesize Speedup} & {\footnotesize Latency (s)} & {\footnotesize Speedup} & {\footnotesize Latency (s)} & {\footnotesize Speedup} \\
\midrule
Official   & 108.3 & \textbf{-} & 118.1 & \textbf{-} & 34.2 & \textbf{-} \\
SGLang     & 99.6 & \textbf{1.00$\times$} & 97.8 & \textbf{1.00$\times$} & 29.4 & \textbf{1.00$\times$} \\
Sol-Engine & 43.9 & \textbf{2.27$\times$} & 41.0 & \textbf{2.38$\times$} & 10.6 & \textbf{2.77$\times$} \\
\bottomrule
\end{tabular}%
}
\end{table}

Across all three deployments, Sol Video Inference Engine achieves more than $2\times$ end-to-end acceleration despite substantial differences in model architecture, parameter scale, and inference pipeline.
This suggests that the agentic acceleration framework is not tied to a single backbone, but can identify and compose effective optimizations for diverse video diffusion systems.
Table~\ref{tab:ablation_stack} summarizes the resulting efficiency gains, with SGLang normalized to $1.00\times$ and official latencies shown for reference.

\paragraph{Cosmos3-Super.}
Cosmos3-Super starts from a 108.3\,s official pipeline and a 99.6\,s SGLang deployment. The diffusion-cache agent explores cache variants for the 64B MoT setting and selects a TeaCache-style policy~\cite{liu2024timestep}, reducing latency from 99.6\,s to 52.4\,s (\textcolor{nvidiagreen}{$1.90\times$} over SGLang).
Quantization is also selective across denoising time: using low precision in early steps can introduce larger structural deviation, while using low precision in the final refinement steps can create high-frequency visual artifacts.
The agent therefore assigns precision by timestep and module sensitivity rather than applying one uniform format everywhere. Kernel-wise optimization reduces latency to 43.9\,s, yielding cumulative \textcolor{nvidiagreen}{$2.27\times$} speedup over SGLang and \textcolor{nvidiagreen}{$2.47\times$} over the official pipeline.

\paragraph{LTX-2.3.}
LTX-2.3 starts from a 118.1\,s official pipeline and a 97.8\,s SGLang deployment. Diffusion cache removes redundancy across repeated denoising steps; because the high-quality LTX pipeline uses the \texttt{res\_2s} sampler, cache schedules originally tuned for Euler-style samplers do not directly fit this trajectory well. In our framework, the agent discovers a fixed-step skip strategy that gives a better speed--quality operating point under the same inference NFE budget, reducing latency from 97.8\,s to 70.3\,s (\textcolor{nvidiagreen}{$1.39\times$} over SGLang).
Sparse attention and token pruning target the high-resolution stage, where the larger token count increases attention and DiT-forward cost; sparse attention reduces unnecessary bidirectional dense attention, while token pruning removes redundant token computation, bringing latency to 58.5\,s (\textcolor{nvidiagreen}{$1.20\times$} incremental and \textcolor{nvidiagreen}{$1.67\times$} cumulative over SGLang).
Kernel fusion then compresses memory-bound and fragmented operators: for example, GELU can be fused as a linear-layer epilogue to avoid a separate kernel launch. Along with NVFP4 quantization, kernel-wise optimization reduces the final latency to 41.0\,s, yielding \textcolor{nvidiagreen}{$2.38\times$} speedup over SGLang and \textcolor{nvidiagreen}{$2.88\times$} over the official pipeline.

\paragraph{SANA-Video.}
SANA-Video starts from a 34.2\,s official pipeline and a 29.4\,s SGLang deployment. SANA-Video already incorporates efficiency-oriented model design, especially linear attention, so model-level acceleration has less room and we focus on the diffusion algorithm and kernel levels.
The cache agent selects EasyCache~\cite{zhou2025easycache} as the best policy for this pipeline, reducing latency from 29.4\,s to 19.8\,s (\textcolor{nvidiagreen}{$1.48\times$} over SGLang).
At the kernel level, replacing high-precision linear-attention kernels with low-bit kernels gives an additional roughly \textcolor{nvidiagreen}{$1.2\times$} kernel-level speedup with very small reconstruction loss (PSNR above 30\,dB), while \texttt{torch.compile} performs operator fusion and graph construction to reduce kernel-launch overhead; together, these changes reduce end-to-end latency to 10.6\,s, yielding \textcolor{nvidiagreen}{$2.77\times$} speedup over SGLang and \textcolor{nvidiagreen}{$3.23\times$} over the official pipeline.

\subsection{Quality Evaluation}
\label{subsec:quality_ltx}

Table~\ref{tab:quality_metrics} compares the unaccelerated baseline against the resulting full-stack accelerated pipeline using detailed VBench dimensions. The reported results demonstrate that the accelerated stacks maintain average evaluation scores on par with the baseline, introducing only negligible variations in visual and motion quality. The evaluation results indicate that the observed latency gains stem from eliminating redundant or inefficient operations rather than sacrificing perceptually critical computations, thereby preserving prompt adherence and overall visual fidelity.

\newcommand{\dposlow}[1]{\cellcolor{green!10}#1}
\newcommand{\dposhigh}[1]{\cellcolor{green!18}#1}
\newcommand{\dneglow}[1]{\cellcolor{red!10}#1}
\newcommand{\dneghigh}[1]{\cellcolor{red!18}#1}

\begin{table}[t]
\centering
\caption{\textbf{VBench quality evaluation under full-stack acceleration.} The table compares the unaccelerated baseline against the fully optimized Sol Video Inference Engine configuration across the retained VBench dimensions. $\Delta$ reports the relative percentage change with respect to the baseline.}
\label{tab:quality_metrics}
\resizebox{\textwidth}{!}{%
\begin{tabular}{lcccccccc}
\toprule
Configuration & \textbf{Average Score} & \makecell{Subject\\consistency} & \makecell{Background\\consistency} & \makecell{Temporal\\flicker} & \makecell{Motion\\smoothness} & \makecell{Aesthetic\\quality} & \makecell{Imaging\\quality} & \makecell{Overall\\consistency} \\
\midrule
\multicolumn{9}{l}{\emph{Cosmos3-Super (64B)}} \\
\cmidrule(lr){1-9}
Baseline & \textbf{0.7759} & 0.9687 & 0.9301 & 0.9859 & 0.9923 & 0.6134 & 0.7276 & 0.2133 \\
Sol-Engine & \textbf{0.7775} & 0.9723 & 0.9382 & 0.9877 & 0.9935 & 0.6197 & 0.7178 & 0.2133 \\
$\Delta$ & {\cellcolor{green!10}\textbf{+0.21\%}} & +0.37\% & +0.87\% & +0.18\% & +0.12\% & +1.03\% & -1.35\% & +0.00\% \\
\midrule
\multicolumn{9}{l}{\emph{LTX-2.3 (22B)}} \\
\cmidrule(lr){1-9}
Baseline & \textbf{0.7646} & 0.9010 & 0.9245 & 0.9675 & 0.9871 & 0.6234 & 0.7012 & 0.2474 \\
Sol-Engine & \textbf{0.7605} & 0.9006 & 0.9137 & 0.9704 & 0.9840 & 0.6104 & 0.7013 & 0.2429 \\
$\Delta$ & {\cellcolor{red!10}\textbf{-0.54\%}} & -0.04\% & -1.17\% & +0.30\% & -0.31\% & -2.09\% & +0.01\% & -1.82\% \\
\midrule
\multicolumn{9}{l}{\emph{SANA-Video (2B)}} \\
\cmidrule(lr){1-9}
Baseline & \textbf{0.7864} & 0.9730 & 0.9648 & 0.9626 & 0.9843 & 0.6650 & 0.6892 & 0.2660 \\
Sol-Engine & \textbf{0.7847} & 0.9750 & 0.9654 & 0.9646 & 0.9842 & 0.6624 & 0.6779 & 0.2637 \\
$\Delta$ & {\cellcolor{red!10}\textbf{-0.21\%}} & +0.21\% & +0.06\% & +0.21\% & -0.01\% & -0.39\% & -1.64\% & -0.86\% \\
\bottomrule
\end{tabular}%
}
\end{table}

\begin{figure}[t]
\centering
\includegraphics[width=\linewidth]{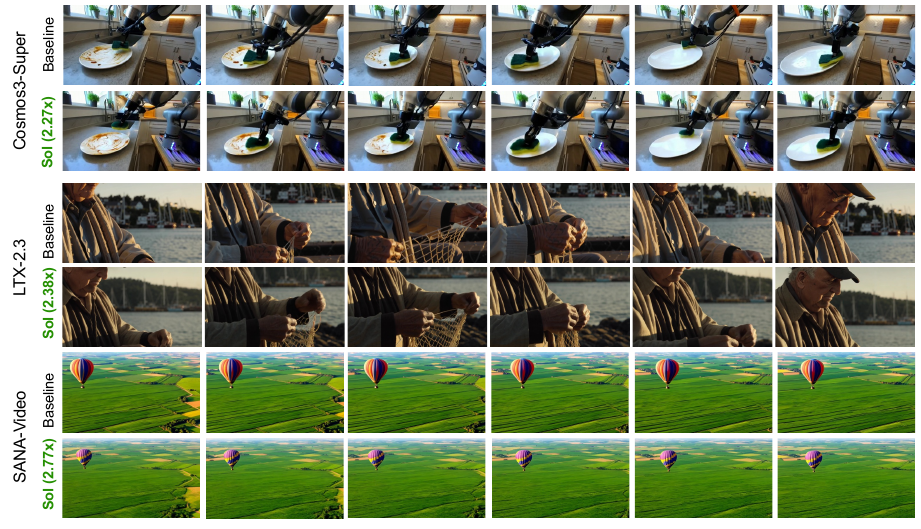}
\caption{\textbf{Visual comparison on generated videos.} Qualitative demo comparing unaccelerated and accelerated outputs under matched prompts and serving configurations, with latency annotations. Our framework achieves substantial speedup without introducing visually perceptible quality differences.}
\label{fig:demo_comparison}
\end{figure}

Figure~\ref{fig:demo_comparison} provides qualitative side-by-side video outputs under matched prompts and serving configurations. The accelerated outputs may exhibit minor differences in fine visual details, but the overall visual quality, motion quality, and physical fidelity remain strong. Across the shown cases, the accelerated videos preserve the basic scene layout, subject arrangement, and motion structure of the unaccelerated baseline, which is the key requirement for deploying the optimized stack in a near-lossless serving regime.
% Fig. 7 — qualitative video demo comparison

\subsection{Extension to NVIDIA B300}
\label{subsec:b300_extension}

\noindent
\begin{minipage}[t]{0.54\textwidth}
\vspace{0pt}
We further evaluate single-GPU Cosmos3-Super on NVIDIA B200 and B300. Relative to B200, B300 maintains similar BF16 compute capability while providing higher NVFP4 compute capability~\cite{nvidia2026hgxcomponents}. Accordingly, LTX-2.3 and SANA-Video, which rely less on NVFP4-heavy execution, show little end-to-end change in our single-GPU setting. By contrast, Cosmos3-Super depends more on NVFP4 execution and therefore obtains a clearer end-to-end speedup on B300. Table~\ref{tab:b300_cosmos_single_gpu} reports single-GPU latency and speedup normalized to the baseline on the same card. On B200, Sol-Engine reduces latency from 353.5\,s to 143.8\,s (\textcolor{nvidiagreen}{$2.46\times$}); on B300, it reduces latency from 351.9\,s to 137.6\,s (\textcolor{nvidiagreen}{$2.56\times$}).
\end{minipage}\hfill
\begin{minipage}[t]{0.43\textwidth}
\vspace{0pt}
\captionsetup{type=table}
\captionof{table}{\textbf{Single-GPU Cosmos3-Super on B200 and B300.} Speedups are normalized to the baseline on the same hardware. It shows a clearer B300 acceleration advantage with the NVFP4 stack.}
\label{tab:b300_cosmos_single_gpu}
\scriptsize
\centering
\resizebox{\linewidth}{!}{%
\begin{tabular}{lcc}
\toprule
\textbf{Framework} & \textbf{Latency (s)} & \textbf{Speedup} \\
\midrule
B200 $\times$ SGLang     & 353.5 & \textbf{1.00$\times$} \\
B200 $\times$ Sol-Engine & 143.8 & \textbf{2.46$\times$} \\
\arrayrulecolor{black!35}\midrule
\arrayrulecolor{black}
B300 $\times$ SGLang     & 351.9 & \textbf{1.00$\times$} \\
B300 $\times$ Sol-Engine & 137.6 & \textbf{2.56$\times$} \\
\bottomrule
\end{tabular}%
}

\end{minipage}

\section{Conclusion}
\label{sec:conclusion}

Modern large-scale video diffusion models are rapidly improving in quality, but their inference cost remains a major obstacle to practical deployment.
We argued that a core systems challenge is \textbf{instance-specific acceleration}: effective speedups depend jointly on model architecture, inference serving configuration, and hardware platform, and recipes tuned for one deployment instance often fail to transfer to another.
As these factors combine into a large optimization space, manually engineering a separate acceleration recipe for every instance requires substantial engineering effort.

Sol Video Inference Engine addresses this gap with an \textbf{agentic acceleration stack} and an \textbf{agent-native acceleration workflow}.
The stack organizes cache, quantization, efficient sparse attention, token pruning, and kernel optimization into model-specific design spaces; parallel skill agents tune local candidates, an agent integrator composes the selected candidates into a deployment stack, and a human validator provides feedback on the final efficiency--quality trade-off for targets including Cosmos3-Super, LTX-2.3, and SANA-Video.
The optimized systems organize these candidates into coherent pipelines tuned per deployment instance.
Across these models, Sol Video Inference Engine achieves more than $2\times$ end-to-end speedup across different model scales and architectural designs while preserving visual and motion quality with negligible degradation.
These results show that agent-native full-stack acceleration can reduce inference cost with little human effort while largely maintaining generation quality.

\paragraph{Limitations and future work.}
While Sol Video Inference Engine demonstrates consistent acceleration across models with different scales, architectures, and serving pipelines, its current workflow still depends on human judgment for final quality assessment.
Traditional automatic loss or similarity operators, such as PSNR-style measurements, are not well aligned with the visual intent of video generation.
When an accelerated output changes harmless visual details, these metrics may report a large difference even though the result is acceptable to a human reviewer.
Conversely, high-frequency noise, blur, temporal jitter, or physically implausible motion may remain numerically close to the baseline while being clearly unacceptable perceptually.
Because these traditional detectors do not faithfully reflect human aesthetic intent and deployment expectations, the current system keeps human feedback validation in the loop: humans inspect representative generations, update the acceptable quality envelope, and redirect the agent search when automatic scores fail to capture visible artifacts.

This limitation also points to a natural direction for future work.
A stronger end-to-end visual-quality evaluator could reduce the amount of manual validation, make the feedback loop more scalable, and allow agents to optimize more aggressively while staying within quality constraints.
Future versions of the framework could incorporate learned video preference models, artifact detectors, physics-consistency checks, and VLM-based end-to-end quality judgment so that the integrator receives richer feedback than latency and aggregate benchmark scores alone.
Improving automatic quality recognition would make the same agent-native acceleration workflow more autonomous and more broadly deployable.

\newpage
\appendix
\onecolumn
\section{Implementation Details}
\label{app:implementation}

\subsection{Overall Implementation Details}

Across all three models, we vary only their concrete implementation scope and hyperparameters: cache policy, sparse attention replacement, token-pruning schedule, low-precision quantization scope, and kernel-level fusion. The deployment interface is always organized into a dense \texttt{baseline} path and a composed \texttt{fullopt} path, so the measured comparisons reflect end-to-end serving stacks rather than isolated kernel microbenchmarks. The runtime itself remains SGLang-based, while the optimized path selectively swaps in cache reuse, sparse or reduced token computation, low-precision kernels, and fused operator sequences according to each model's architecture and bottlenecks.

At a method level, the global implementation principles are consistent. Cache is always applied as a training-free denoising-step reuse policy, but the exact schedule differs by model. Sparse attention is only introduced when the attention map exhibits enough structure to justify replacing dense bidirectional attention with a piecewise sparse pattern. Token pruning is used only when intermediate video tokens contain enough redundancy that dropping a subset can reduce cost without visible artifacts. Quantization is applied selectively instead of uniformly: numerically fragile stages stay in higher precision, while stable GEMM-heavy regions are moved to low precision. Kernel optimization is similarly selective and includes epilogue fusion, QKV-path fusion, normalization-path fusion, and compiler-driven graph fusion when those transformations are stable for the target workload.

\subsection{Model-Specific Pipelines}

\subsubsection{Cosmos3-Super}

Cosmos3-Super is dominated by very large transformer blocks, so the optimized pipeline focuses on reducing repeated denoising computation and accelerating the GEMM-heavy middle part of the trajectory. The cache component follows a TeaCache-style residual replay policy with threshold 1.15, start step 10, and at most three continuous cache hits. It measures the relative $L_1$ change of the generation hidden state across denoising steps, accumulates this change, and skips the generation-path transformer blocks only when the accumulated change remains below the threshold; otherwise it recomputes and refreshes the cached residual. Quantization is applied in a step-selective manner rather than uniformly. The first three and last three denoising steps are kept in the dense/high-precision path, because early steps determine global structure while late steps refine high-frequency details. NVFP4 is then applied to the more regular middle steps and GEMM-heavy generation-path linear layers with high arithmetic density, including the FFN gate-up and down projections as well as the attention QKV and output projections. This combination makes Cosmos3-Super the case where cache and NVFP4-oriented transformer execution provide the clearest end-to-end gain.

\subsubsection{LTX-2.3}

LTX-2.3 uses a two-stage high-quality pipeline. Stage~1 runs the \texttt{res\_2s} sampler for 15 denoising steps at $544{\times}960$, and stage~2 upsamples and refines the latent at $1088{\times}1920$ for 3 denoising steps before VAE decoding 241 frames. The full optimized stack uses a fixed-step cache method in stage~1 rather than an online TeaCache threshold. The concrete preset is \texttt{8of15\_last\_29calls}: over the 29 stage-1 denoiser calls, it reuses the last computed result at call indices [13, 14, 16, 17, 18, 19, 20, 21, 22, 24, 25, 26, 27], with \texttt{reuse\_mode=last} and \texttt{max\_skip\_steps=30}. Stage~2 applies PISA only to the high-resolution refinement transformer: sparsity is 0.9 (density 0.1), block size is 64, only video self-attention is replaced, stage-1 sparse scheduling is disabled, and stage-2 layers 0--1 remain dense while later layers use piecewise attention with approximate remainder enabled, score-based routing, and FlashAttention dense fallback. Token pruning is also restricted to stage~2: during the additional refinement model calls, the system keeps 50\% of video tokens according to the \texttt{feat\_norm} saliency score, and applies pruning on refinement calls 1 and 2. NVFP4 is applied only to the video FFN input and output projections; the NVFP4 fused \texttt{proj\_in+GELU} and \texttt{proj\_out+bias+gate} paths are disabled. The lossless kernel/fusion flags enabled in the full stack are block-0 self-attention sharing, guidance-prefix sharing, fused QK+RoPE, fused RMS-AdaLN, fused AdaLN, fused QKNorm+RoPE, fused dual modulation, fused cross-attention dual modulation, fused Ada values, fused residual gate, fused FFN \texttt{proj\_in+GELU}, compiled gate-to-output, fused audio QKVG, global fused QKNorm+RoPE, and compiled tiled VAE decoding. The distilled LoRA strengths are 0.25 in stage~1 and 0.5 in stage~2; stage~2 uses sigmas $[0.909375, 0.725, 0.421875, 0.0]$.

\subsubsection{SANA-Video}

SANA-Video uses a 2B linear-attention backbone and is evaluated with 81 frames, 50 denoising steps, and the 480p setting $832{\times}480$ in the main demo configuration. Its optimized path does not use sparse attention, token pruning, or NVFP4 FFN quantization. The cache component is EasyCache with threshold 0.1, warmup 3 steps, and spatial subsample stride 8. The decision rule estimates the online input-to-output transformation rate from subsampled hidden states, accumulates the estimated relative output change, and skips the transformer block stack while the accumulated value stays below the threshold; the cached residual is shared across the unbatched CFG branches to avoid injecting stale guidance. The remaining acceleration comes from three lossless or near-lossless kernel-path changes: the linear-attention key/value aggregation is kept in BF16 tensor-core execution instead of being promoted to FP32, self-attention Q/K/V projections are merged into one GEMM by concatenating the projection weights after loading, and the DiT block stack is compiled with torch.compile. The max-autotune compile mode uses \texttt{max-autotune-no-cudagraphs}, subprocess autotuning, and a persistent TorchInductor cache; the safe compile path uses the default torch.compile mode. The 480p setting reports roughly 16 skipped steps out of 50 for EasyCache, with the remaining speedup coming from BF16 linear attention, QKV merge, and compile/fusion.

\section{Visual Demo Comparison}
\label{app:visual_demos}

We provide extended visual comparisons to show that the Sol-Engine accelerated pipelines remain visually close to the unaccelerated SGLang deployments while achieving more than $2\times$ end-to-end speedup. For each of Cosmos3-Super, LTX-2.3, and SANA-Video, we collect six matched video pairs. In each figure, the upper strip shows the unaccelerated SGLang result and the lower strip shows the Sol-Engine result for the same prompt, making it possible to inspect whether scene layout, motion evolution, and visual fidelity are preserved under the accelerated setting.

\begin{figure}[p]
\centering
\includegraphics[width=\linewidth,height=0.86\textheight,keepaspectratio]{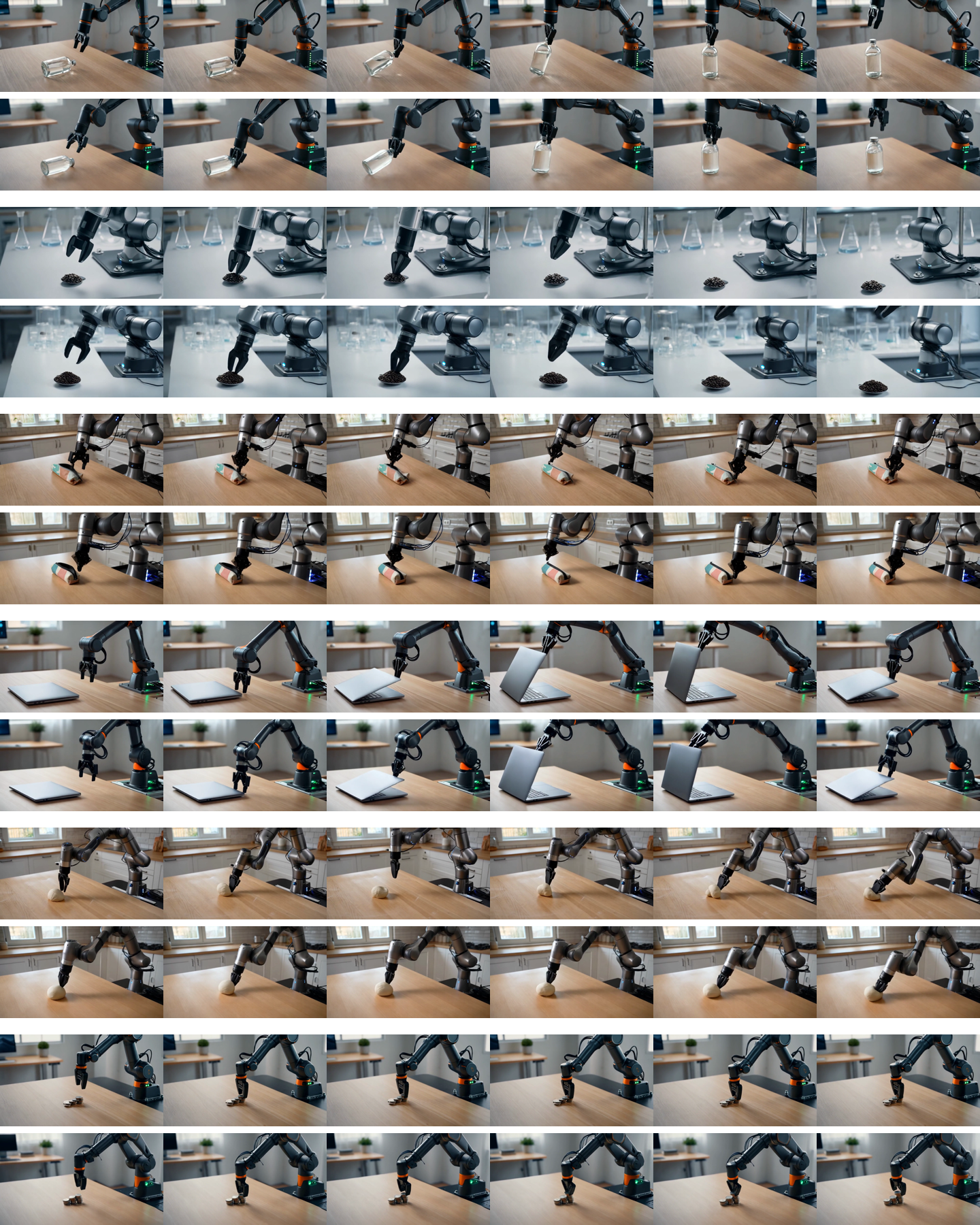}
\caption{\textbf{Cosmos3-Super visual demo comparisons.} The video pairs compare the unaccelerated SGLang deployment with the Sol-Engine accelerated pipeline. The upper strip shows the unaccelerated SGLang result and the lower strip shows the Sol-Engine result. With $2.27\times$ end-to-end acceleration over SGLang, the accelerated pipeline preserves the main scene layout, object structure, and motion evolution.}
\label{fig:appendix_cosmos_demos}
\end{figure}

\begin{figure}[p]
\centering
\includegraphics[width=\linewidth,height=0.86\textheight,keepaspectratio]{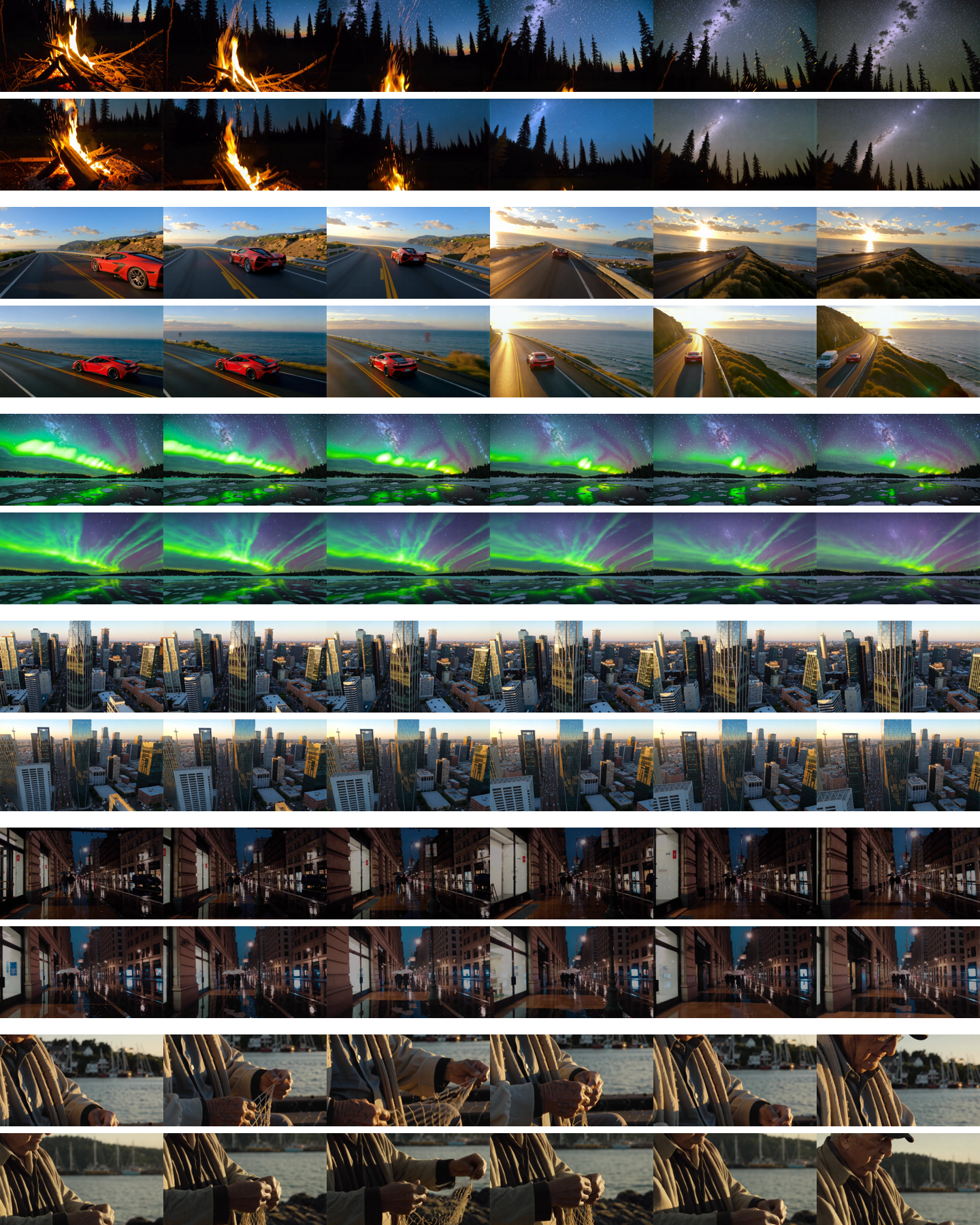}
\caption{\textbf{LTX-2.3 visual demo comparisons.} The video pairs compare the unaccelerated SGLang deployment with the Sol-Engine accelerated pipeline. The upper strip shows the unaccelerated SGLang result and the lower strip shows the Sol-Engine result. With $2.38\times$ end-to-end acceleration over SGLang, the accelerated pipeline maintains the overall visual quality, temporal coherence, and scene composition.}
\label{fig:appendix_ltx_demos}
\end{figure}

\begin{figure}[p]
\centering
\includegraphics[width=\linewidth,height=0.86\textheight,keepaspectratio]{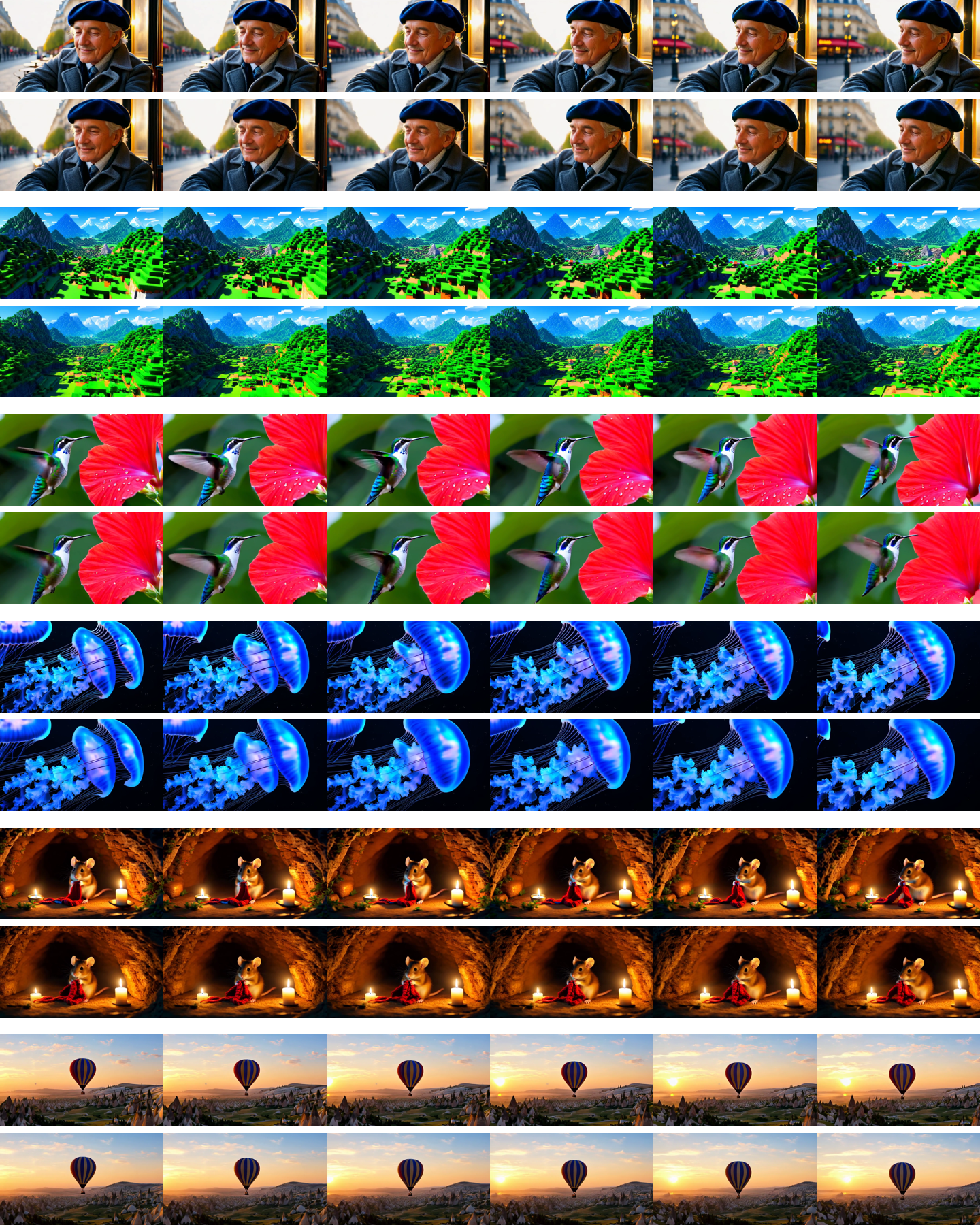}
\caption{\textbf{SANA-Video visual demo comparisons.} The video pairs compare the unaccelerated SGLang deployment with the Sol-Engine accelerated pipeline. The upper strip shows the unaccelerated SGLang result and the lower strip shows the Sol-Engine result. With $2.77\times$ end-to-end acceleration over SGLang, the accelerated pipeline preserves prompt-level appearance, motion trends, and perceptual quality.}
\label{fig:appendix_sana_demos}
\end{figure}

\clearpage
{
  \small
  \bibliographystyle{unsrtnat}
  \bibliography{ref}

@article{liu2024timestep,
  title={Timestep Embedding Tells: It's Time to Cache for Video Diffusion Model},
  author={Liu, Feng and Zhang, Shiwei and Wang, Xiaofeng and Wei, Yujie and Qiu, Haonan and Zhao, Yuzhong and Zhang, Yingya and Ye, Qixiang and Wan, Fang},
  journal={arXiv preprint arXiv:2411.19108},
  year={2024}
}

@article{zhou2025easycache,
  title={Less is Enough: Training-Free Video Diffusion Acceleration via Runtime-Adaptive Caching},
  author={Zhou, Xin and Liang, Dingkang and Chen, Kaijin and Feng, Tianrui and Chen, Xiwu and Lin, Hongkai and Ding, Yikang and Tan, Feiyang and Zhao, Hengshuang and Bai, Xiang},
  journal={arXiv preprint arXiv:2507.02860},
  year={2025}
}

@misc{cachedit2025,
  title={Cache-DiT: A PyTorch-native Inference Engine with Cache, Parallelism and Quantization for Diffusion Transformers},
  author={{DefTruth, vipshop.com, etc.}},
  howpublished={\url{https://github.com/vipshop/cache-dit.git}},
  note={Open-source software. Accessed June 20, 2026},
  year={2025}
}

@article{zhao2024pab,
  title={Real-Time Video Generation with Pyramid Attention Broadcast},
  author={Zhao, Xuanlei and Jin, Xiaolong and Wang, Kai and You, Yang},
  journal={arXiv preprint arXiv:2408.12588},
  year={2024}
}

@inproceedings{liu2025taylorseer,
  title={From Reusing to Forecasting: Accelerating Diffusion Models with TaylorSeers},
  author={Liu, Jiacheng and Zou, Chang and Lyu, Yuanhuiyi and Chen, Junjie and Zhang, Linfeng},
  booktitle={Proceedings of the IEEE/CVF International Conference on Computer Vision (ICCV)},
  month={October},
  year={2025},
  pages={15853--15863}
}

@article{xi2025sparse,
  title={Sparse VideoGen: Accelerating Video Diffusion Transformers with Spatial-Temporal Sparsity},
  author={Xi, Haocheng and Yang, Shuo and Zhao, Yilong and Xu, Chenfeng and Li, Muyang and Li, Xiuyu and Lin, Yujun and Cai, Han and Zhang, Jintao and Li, Dacheng and others},
  journal={arXiv preprint arXiv:2502.01776},
  year={2025}
}

@article{yang2025sparse,
  title={Sparse VideoGen2: Accelerate Video Generation with Sparse Attention via Semantic-Aware Permutation},
  author={Yang, Shuo and Xi, Haocheng and Zhao, Yilong and Li, Muyang and Zhang, Jintao and Cai, Han and Lin, Yujun and Li, Xiuyu and Xu, Chenfeng and Peng, Kelly and others},
  journal={arXiv preprint arXiv:2505.18875},
  year={2025}
}

@inproceedings{zhang2025vsa,
  title={Faster Video Diffusion with Trainable Sparse Attention},
  author={Zhang, Peiyuan and Chen, Yongqi and Huang, Haofeng and Lin, Will and Liu, Zhengzhong and Stoica, Ion and Xing, Eric P. and Zhang, Hao},
  booktitle={Advances in Neural Information Processing Systems},
  year={2025}
}

@inproceedings{zhang2025sageattention,
  title={SageAttention: Accurate 8-Bit Attention for Plug-and-Play Inference Acceleration},
  author={Zhang, Jintao and Wei, Jia and Huang, Haofeng and Zhang, Pengle and Zhu, Jun and Chen, Jianfei},
  booktitle={International Conference on Learning Representations},
  year={2025}
}

@inproceedings{zhang2025sageattention2,
  title={SageAttention2: Efficient Attention with Thorough Outlier Smoothing and Per-thread INT4 Quantization},
  author={Zhang, Jintao and Huang, Haofeng and Zhang, Pengle and Wei, Jia and Zhu, Jun and Chen, Jianfei},
  booktitle={Proceedings of the 42nd International Conference on Machine Learning},
  series={Proceedings of Machine Learning Research},
  volume={267},
  pages={75097--75119},
  year={2025}
}

@article{zhang2025sageattention3,
  title={SageAttention3: Microscaling FP4 Attention for Inference and An Exploration of 8-Bit Training},
  author={Zhang, Jintao and Wei, Jia and Wang, Haoxu and Zhang, Pengle and Xu, Xiaoming and Huang, Haofeng and Jiang, Kai and Zhu, Jun and Chen, Jianfei},
  journal={arXiv preprint arXiv:2505.11594},
  year={2025}
}

@misc{li2026fp4explorebf16train,
  title={FP4 Explore, BF16 Train: Diffusion Reinforcement Learning via Efficient Rollout Scaling},
  author={Yitong Li and Junsong Chen and Shuchen Xue and Pengcuo Zeren and Siyuan Fu and Dinghao Yang and Yangyang Tang and Junjie Bai and Ping Luo and Song Han and Enze Xie},
  year={2026},
  eprint={2604.06916},
  archivePrefix={arXiv},
  primaryClass={cs.LG},
  url={https://arxiv.org/abs/2604.06916}
}

@article{li2026pisa,
  title={PISA: Piecewise Sparse Attention Is Wiser for Efficient Diffusion Transformers},
  author={Li, Haopeng and Shao, Shitong and Zhong, Wenliang and Zhou, Zikai and Bai, Lichen and Xiong, Hui and Xie, Zeke},
  journal={arXiv preprint arXiv:2602.01077},
  year={2026}
}

@inproceedings{zhang2025spargeattn,
  title={Spargeattn: Accurate sparse attention accelerating any model inference},
  author={Zhang, Jintao and Xiang, Chendong and Huang, Haofeng and Wei, Jia and Xi, Haocheng and Zhu, Jun and Chen, Jianfei},
  booktitle={International Conference on Machine Learning (ICML)},
  year={2025}
}

@article{shen2025draft,
  title={DraftAttention: Fast Video Diffusion via Low-Resolution Attention Guidance},
  author={Shen, Xuan and Han, Chenxia and Zhou, Yufa and Xie, Yanyue and Gong, Yifan and Wang, Quanyi and Wang, Yiwei and Wang, Yanzhi and Zhao, Pu and Gu, Jiuxiang},
  journal={arXiv preprint arXiv:2505.14708},
  year={2025}
}

@inproceedings{zhang2025sta,
  title={Fast Video Generation with Sliding Tile Attention},
  author={Zhang, Peiyuan and Chen, Yongqi and Su, Runlong and Ding, Hangliang and Stoica, Ion and Liu, Zhengzhong and Zhang, Hao},
  booktitle={Proceedings of the 42nd International Conference on Machine Learning},
  series={Proceedings of Machine Learning Research},
  volume={267},
  pages={74714--74731},
  year={2025},
  url={https://proceedings.mlr.press/v267/zhang25m.html}
}

@article{bolya2023tomesd,
  title={Token Merging for Fast Stable Diffusion},
  author={Bolya, Daniel and Hoffman, Judy},
  journal={CVPR Workshop on Efficient Deep Learning for Computer Vision},
  year={2023}
}

@article{liu2025astraea,
  title={Astraea: A Token-wise Acceleration Framework for Video Diffusion Transformers},
  author={Liu, Haosong and Cheng, Yuge and Miao, Wenxuan and Liu, Zihan and Chen, Aiyue and Lin, Jing and Yao, Yiwu and Chen, Chen and Leng, Jingwen and Feng, Yu and Guo, Minyi},
  journal={arXiv preprint arXiv:2506.05096},
  year={2025}
}

@article{li2026tape,
  title={Temporal Aware Pruning for Efficient Diffusion-based Video Generation},
  author={Li, Sheng and Sui, Yang and Ran, Junhao and Yuan, Bo and Dai, Yue and Tang, Xulong},
  journal={arXiv preprint arXiv:2605.17837},
  year={2026}
}

@article{li2026coredit,
  title={CoReDiT: Spatial Coherence-Guided Token Pruning and Reconstruction for Efficient Diffusion Transformers},
  author={Li, Zhuojin and Cheng, Hsin-Pai and Cai, Hong and Han, Shizhong and Porikli, Fatih},
  journal={arXiv preprint arXiv:2605.14191},
  year={2026}
}

@misc{cutlassEpilogue,
  title={CUTLASS Epilogue Operations},
  author={{NVIDIA}},
  howpublished={\url{https://nvidia-cutlass-22.mintlify.app/cpp/epilogue}},
  note={Documentation. Accessed June 20, 2026},
  year={2025}
}

@inproceedings{zhai2023bytetransformer,
  title={ByteTransformer: A High-Performance Transformer Boosted for Variable-Length Inputs},
  author={Zhai, Yujia and Jiang, Chengquan and Wang, Leyuan and Jia, Xiaoying and Zhang, Shang and Chen, Zizhong and Liu, Xin and Zhu, Yibo},
  booktitle={2023 IEEE International Parallel and Distributed Processing Symposium (IPDPS)},
  pages={344--355},
  year={2023}
}

@article{guo2026coda,
  title={CODA: Rewriting Transformer Blocks as GEMM-Epilogue Programs},
  author={Guo, Han and Zhang, Jack and Menon, Arjun and Guessous, Driss and Thakkar, Vijay and Kim, Yoon and Dao, Tri},
  journal={arXiv preprint arXiv:2605.19269},
  year={2026}
}

@inproceedings{wu2024ptq4dit,
  title={PTQ4DiT: Post-training Quantization for Diffusion Transformers},
  author={Wu, Junyi and Wang, Haoxuan and Shang, Yuzhang and Shah, Mubarak and Yan, Yan},
  booktitle={Advances in Neural Information Processing Systems},
  year={2024}
}

@inproceedings{chen2025qdit,
  title={Q-DiT: Accurate Post-Training Quantization for Diffusion Transformers},
  author={Chen, Lei and Meng, Yuan and Tang, Chen and Ma, Xinzhu and Jiang, Jingyan and Wang, Xin and Wang, Zhi and Zhu, Wenwu},
  booktitle={Proceedings of the IEEE/CVF Conference on Computer Vision and Pattern Recognition (CVPR)},
  pages={28306--28315},
  year={2025}
}

@inproceedings{zhao2025viditq,
  title={ViDiT-Q: Efficient and Accurate Quantization of Diffusion Transformers for Image and Video Generation},
  author={Zhao, Tianchen and Fang, Tongcheng and Huang, Haofeng and Wan, Rui and Soedarmadji, Widyadewi and Liu, Enshu and Li, Shiyao and Lin, Zinan and Dai, Guohao and Yan, Shengen and Yang, Huazhong and Ning, Xuefei and Wang, Yu},
  booktitle={International Conference on Learning Representations},
  year={2025}
}

@inproceedings{li2025svdquant,
  title={SVDQuant: Absorbing Outliers by Low-Rank Component for 4-Bit Diffusion Models},
  author={Li, Muyang and Lin, Yujun and Zhang, Zhekai and Cai, Tianle and Li, Xiuyu and Guo, Junxian and Xie, Enze and Meng, Chenlin and Zhu, Jun-Yan and Han, Song},
  booktitle={International Conference on Learning Representations},
  year={2025}
}

@article{li2025radial,
  title={Radial Attention: $\mathcal{O}(n\log n)$ Sparse Attention with Energy Decay for Long Video Generation},
  author={Li, Xingyang and Li, Muyang and Cai, Tianle and Xi, Haocheng and Yang, Shuo and Lin, Yujun and Zhang, Lvmin and Yang, Songlin and Hu, Jinbo and Peng, Kelly and Agrawala, Maneesh and Stoica, Ion and Keutzer, Kurt and Han, Song},
  journal={arXiv preprint arXiv:2506.19852},
  year={2025}
}

@misc{chen2026longlive20,
  title={{LongLive-2.0}: An NVFP4 Parallel Infrastructure for Long Video Generation},
  author={Chen, Yukang and Wang, Luozhou and Huang, Wei and Yang, Shuai and Zhang, Bohan and Xiao, Yicheng and Chu, Ruihang and Mao, Weian and Hu, Qixin and Liu, Shaoteng and Zhao, Yuyang and Mao, Huizi and Chen, Ying-Cong and Xie, Enze and Qi, Xiaojuan and Han, Song},
  year={2026},
  eprint={2605.18739},
  archivePrefix={arXiv},
  primaryClass={cs.CV},
  url={https://arxiv.org/abs/2605.18739}
}

@inproceedings{xu2025xattention, 
  title = {XAttention: Block Sparse Attention with Antidiagonal Scoring}, 
  author = {Xu, Ruyi and Xiao, Guangxuan and Huang, Haofeng and Guo, Junxian and Han, Song}, 
  booktitle = {Proceedings of the 42nd International Conference on Machine Learning (ICML)}, 
  year = {2025}
}

@article{wan2025,
  title={Wan: Open and Advanced Large-Scale Video Generative Models},
  author={Team Wan and Wang, Ang and Ai, Baole and Wen, Bin and Mao, Chaojie and Xie, Chen-Wei and Chen, Di and Yu, Feiwu and Zhao, Haiming and Yang, Jianxiao and others},
  journal={arXiv preprint arXiv:2503.20314},
  year={2025}
}

@article{kong2024hunyuanvideo,
  title={HunyuanVideo: A Systematic Framework for Large Video Generative Models},
  author={Kong, Weijie and Tian, Qi and Zhang, Zijian and Min, Rox and Dai, Zuozhuo and Zhou, Jin and Xiong, Jiangfeng and Li, Xin and Wu, Bo and Zhang, Jianwei and others},
  journal={arXiv preprint arXiv:2412.03603},
  year={2024}
}

@misc{hunyuanvideo2025,
      title={HunyuanVideo 1.5 Technical Report}, 
      author={Tencent Hunyuan Foundation Model Team},
      year={2025},
      eprint={2511.18870},
      archivePrefix={arXiv},
      primaryClass={cs.CV},
      url={https://arxiv.org/abs/2511.18870}, 
}

@article{jin2024pyramidal,
  title={Pyramidal Flow Matching for Efficient Video Generative Modeling},
  author={Jin, Yang and Sun, Zhicheng and Li, Ningyuan and Xu, Kun and Xu, Kun and Jiang, Hao and Zhuang, Nan and Huang, Quzhe and Song, Yang and Mu, Yadong and Lin, Zhouchen},
  journal={arXiv preprint arXiv:2410.05954},
  year={2024}
}

@techreport{echo2026longvideo,
  title        = {JoyAI-Echo: Pushing the Frontier of Long Video Generation},
  author       = {{Echo Team @ Joy Future Academy, JD}},
  institution  = {Joy Future Academy, JD},
  year         = {2026},
  month        = {May},
  url          = {https://echo-team-joy-future-academy-jd.github.io/Echo-LongVideo-Page/},
  note         = {Project page. Accessed June 20, 2026}
}

@article{hong2022cogvideo,
  title={CogVideo: Large-scale Pretraining for Text-to-Video Generation via Transformers},
  author={Hong, Wenyi and Ding, Ming and Zheng, Wendi and Liu, Xinghan and Tang, Jie},
  journal={arXiv preprint arXiv:2205.15868},
  year={2022}
}

@inproceedings{yang2025cogvideox,
  title={CogVideoX: Text-to-Video Diffusion Models with an Expert Transformer},
  author={Yang, Zhuoyi and Teng, Jiayan and Zheng, Wendi and Ding, Ming and Huang, Shiyu and Xu, Jiazheng and Yang, Yuanming and Hong, Wenyi and Zhang, Xiaohan and Feng, Guanyu and Yin, Da and Zhang, Yuxuan and Wang, Weihan and Cheng, Yean and Xu, Bin and Gu, Xiaotao and Dong, Yuxiao and Tang, Jie},
  booktitle={International Conference on Learning Representations},
  year={2025}
}

@misc{meituan2025longcatvideo,
  title={LongCat-Video Technical Report},
  author={Meituan LongCat Team and Cai, Xunliang and Huang, Qilong and Kang, Zhuoliang and Li, Hongyu and Liang, Shijun and Ma, Liya and Ren, Siyu and Wei, Xiaoming and Xie, Rixu and Zhang, Tong},
  year={2025},
  eprint={2510.22200},
  archivePrefix={arXiv},
  primaryClass={cs.CV},
  url={https://arxiv.org/abs/2510.22200}
}

@misc{lightricks2026ltx23,
  title={{LTX-2.3 Model Card}},
  author={{Lightricks}},
  howpublished={\url{https://huggingface.co/Lightricks/LTX-2.3}},
  note={Model checkpoint family including ltx-2.3-22b-dev and distilled variants. Accessed June 20, 2026},
  year={2026}
}

@article{nvidia2026cosmos3super,
  title={{Cosmos 3}: Omnimodal World Models for Physical AI},
  author={{NVIDIA}},
  journal={arXiv preprint arXiv:2606.02800},
  url={https://arxiv.org/abs/2606.02800},
  year={2026}
}

@misc{chen2025sana,
  title={{SANA-Video}: Efficient Video Generation with Block Linear Diffusion Transformer},
  author={Chen, Junsong and Zhao, Yuyang and Yu, Jincheng and Chu, Ruihang and Chen, Junyu and Yang, Shuai and Wang, Xianbang and Pan, Yicheng and Zhou, Daquan and Ling, Huan and others},
  year={2025},
  eprint={2509.24695},
  archivePrefix={arXiv},
  primaryClass={cs.CV},
  url={https://arxiv.org/abs/2509.24695}
}

@misc{nvidia2026hgxcomponents,
  title={Components --- NVIDIA HGX AI Factory},
  author={{NVIDIA}},
  howpublished={\url{https://docs.nvidia.com/enterprise-reference-architectures/hgx-ai-factory/latest/components.html}},
  note={Accessed June 20, 2026},
  year={2026}
}

@inproceedings{huang2024vbench,
  title={VBench: Comprehensive Benchmark Suite for Video Generative Models},
  author={Huang, Ziqi and He, Yinan and Yu, Jiashuo and Zhang, Fan and Si, Chenyang and Jiang, Yuming and Zhang, Yuanhan and Wu, Tianxing and Jin, Qingyang and Chanpaisit, Nattapol and Wang, Yaohui and Chen, Xinyuan and Wang, Limin and Lin, Dahua and Qiao, Yu and Liu, Ziwei},
  booktitle={Proceedings of the IEEE/CVF Conference on Computer Vision and Pattern Recognition},
  pages={21807--21818},
  year={2024}
}

@article{zhong2026ahe,
  title={AI Harness Engineering: A Runtime Substrate for Foundation-Model Software Agents},
  author={Zhong, Hailin and Zhu, Shengxin},
  journal={arXiv preprint arXiv:2605.13357},
  year={2026}
}

@inproceedings{huang2024mlagentbench,
  title={MLAgentBench: Evaluating Language Agents on Machine Learning Experimentation},
  author={Huang, Qian and Vora, Jian and Liang, Percy and Leskovec, Jure},
  booktitle={Proceedings of the 41st International Conference on Machine Learning},
  series={Proceedings of Machine Learning Research},
  volume={235},
  pages={20271--20309},
  year={2024}
}

@inproceedings{liu2024agentbench,
  title={AgentBench: Evaluating LLMs as Agents},
  author={Liu, Xiao and Yu, Hao and Zhang, Hanchen and Xu, Yifan and Lei, Xuanyu and Lai, Hanyu and Gu, Yu and Ding, Hangliang and Men, Kaiwen and Yang, Kejuan and Zhang, Shudan and Deng, Xiang and Zeng, Aohan and Du, Zhengxiao and Zhang, Chenhui and Shen, Sheng and Zhang, Tianjun and Su, Yu and Sun, Huan and Huang, Minlie and Dong, Yuxiao and Tang, Jie},
  booktitle={International Conference on Learning Representations},
  year={2024}
}

@inproceedings{yang2024sweagent,
  title={SWE-agent: Agent-Computer Interfaces Enable Automated Software Engineering},
  author={Yang, John and Jimenez, Carlos E. and Wettig, Alexander and Lieret, Kilian and Yao, Shunyu and Narasimhan, Karthik R. and Press, Ofir},
  booktitle={Advances in Neural Information Processing Systems},
  year={2024}
}

@article{zhang2024autocoderover,
  title={AutoCodeRover: Autonomous Program Improvement},
  author={Zhang, Yuntong and Ruan, Haifeng and Fan, Zhiyu and Roychoudhury, Abhik},
  journal={arXiv preprint arXiv:2404.05427},
  year={2024}
}

@article{xia2024agentless,
  title={Agentless: Demystifying LLM-based Software Engineering Agents},
  author={Xia, Chunqiu Steven and Deng, Yinlin and Dunn, Soren and Zhang, Lingming},
  journal={arXiv preprint arXiv:2407.01489},
  year={2024}
}

@article{wang2024openhands,
  title={OpenHands: An Open Platform for AI Software Developers as Generalist Agents},
  author={Wang, Xingyao and Li, Boxuan and Song, Yufan and Xu, Frank F. and Tang, Xiangru and Zhuge, Mingchen and Pan, Jiayi and Song, Yueqi and Li, Bowen and Singh, Jaskirat and Tran, Hoang H. and Li, Fuqiang and Ma, Ren and Zheng, Mingzhang and Qian, Bill and Shao, Yanjun and Muennighoff, Niklas and Zhang, Yizhe and Hui, Binyuan and Lin, Junyang and Brennan, Robert and Peng, Hao and Ji, Heng and Neubig, Graham},
  journal={arXiv preprint arXiv:2407.16741},
  year={2024}
}

@article{lu2024aiscientist,
  title={The AI Scientist: Towards Fully Automated Open-Ended Scientific Discovery},
  author={Lu, Chris and Lu, Cong and Lange, Robert Tjarko and Foerster, Jakob and Clune, Jeff and Ha, David},
  journal={arXiv preprint arXiv:2408.06292},
  year={2024}
}

@article{chen2025cudallm,
  title={CUDA-LLM: LLMs Can Write Efficient CUDA Kernels},
  author={Chen, Wentao and Zhu, Jiace and Fan, Qi and Ma, Yehan and Zou, An},
  journal={arXiv preprint arXiv:2506.09092},
  year={2025}
}

@article{zhang2025cudaforge,
  title={CudaForge: An Agent Framework with Hardware Feedback for CUDA Kernel Optimization},
  author={Zhang, Zijian and Wang, Rong and Li, Shiyang and Luo, Yuebo and Hong, Mingyi and Ding, Caiwen},
  journal={arXiv preprint arXiv:2511.01884},
  year={2025}
}
}

\end{document}